\documentclass{article} 
\usepackage{iclr2027_conference,times}
\iclrfinalcopy
\usepackage{etoolbox}
\makeatletter
\patchcmd{\@maketitle}
  {\lhead{Published as a conference paper at ICLR 2027}}
  {\lhead{}}
  {}{\PackageWarning{arxiv}{Could not remove ICLR header}}
\makeatother

\usepackage{amsmath,amsfonts,bm}

\def\eqref#1{equation~\ref{#1}}

\def\1{\bm{1}}

\def\ve{{\bm{e}}}

\def\vu{{\bm{u}}}

\def\vx{{\bm{x}}}
\def\vy{{\bm{y}}}

\def\mA{{\bm{A}}}
\def\mB{{\bm{B}}}
\def\mC{{\bm{C}}}

\def\mK{{\bm{K}}}
\def\mL{{\bm{L}}}

\DeclareMathAlphabet{\mathsfit}{\encodingdefault}{\sfdefault}{m}{sl}
\SetMathAlphabet{\mathsfit}{bold}{\encodingdefault}{\sfdefault}{bx}{n}

\usepackage{hyperref}
\usepackage{url}

\usepackage[table]{xcolor}         
\usepackage{makecell}
\usepackage{graphicx}
\usepackage{amsthm}
\usepackage{amsmath}
\usepackage{multirow}
\usepackage{float}
\usepackage{caption}
\usepackage{tikz}
\tikzset{>=latex}
\usepackage{booktabs}

\usepackage{xcolor} 
\usepackage{soul}

\usepackage{amssymb}
\usepackage{pifont}
\usepackage{booktabs}
\usepackage{longtable}
\usepackage{pdflscape}

\usepackage{xfp}

\definecolor{resultblue}{RGB}{0,90,180}
\definecolor{resultred}{RGB}{190,30,45}

\newcommand{\result}[2]{%
    \ifnum\pdfstrcmp{\fpeval{#1-#2 > 0}}{1}=0
        {\color{resultblue}\(\mathbf{#1 \pm #2}\)}%
    \else
        \ifnum\pdfstrcmp{\fpeval{#1+#2 < 0}}{1}=0
            {\color{resultred}\(#1 \pm #2\)}%
        \else
            \(#1 \pm #2\)%
        \fi
    \fi
}

\theoremstyle{plain}

\newtheorem{proposition}{Proposition}[section]

\theoremstyle{definition}

\newtheorem{hypothesis}{Hypothesis}[section]
\theoremstyle{remark}

\usepackage{tikz}
\usetikzlibrary{arrows.meta,calc,positioning}

\definecolor{obs}{RGB}{230,105,20}
\definecolor{pred}{RGB}{35,100,220}
\definecolor{outc}{RGB}{20,145,135}
\definecolor{osbg}{RGB}{226,238,249}
\definecolor{ssbg}{RGB}{222,239,236}

\title{Correct then Forecast: Observer State-Space Models for Time Series Forecasting}

\author{%
Alexis-Raja Brachet$^{1,2}$, Guillaume Clavier-Fremond$^1$, Abdelhakim Ziani$^{1,3}$ \\
\textbf{Pierre-Yves Richard}$^{2}$ \& \textbf{Céline Hudelot}$^1$ \vspace{0.2cm} \\ 
$^1$MICS, CentraleSupélec, Université Paris-Saclay, France \\ 
$^2$CentraleSupélec, IETR UMR CNRS 6164, France \\
$^3$Università di Torino, Torino, Italy. \\
\texttt{\{alexisraja.brachet,hakim.ziani\}@centralesupelec.fr} \\
\texttt{guillaume.clavier-fremond@student-cs.fr} \\
\texttt{\{pierre-yves.richard,celine.hudelot\}@centralesupelec.fr}
}

\begin{document}

\maketitle

\begin{abstract}

Time series forecasting requires extrapolating the dynamics of an observed process beyond the last available measurement. Yet recurrent forecasting models typically treat observations as inputs that directly control their latent dynamics. It leads to a regime change when these observations become unavailable at prediction time. Following a state-estimation perspective, we introduce \textbf{Observer State-Space Models} (OSSMs), a class of recurrent models that \textbf{interprets the observed input time series as measurements of an underlying autonomous dynamical system}. OSSMs explicitly separate latent-state propagation from measurement assimilation: a single transition governs the dynamics across both context and forecasting intervals, while available observations \textbf{correct} the estimated state through an \textbf{observer}. This formulation naturally exposes classical control-theoretic properties, including observability and convergence of the state estimation error. We further show that conventional and recent SSMs can be recovered as particular instances of our OSSM framework, thereby providing a unified interpretation of their recurrent dynamics and revealing modeling inconsistencies. We perform experiments across several benchmarks showing that OSSM achieves substantial improvements while maintaining the same parameter count and training setup as the corresponding SSM baseline.  These results support a simple principle for recurrent forecasting: observations should correct the estimated latent state, rather than control the dynamics used to propagate it.
\end{abstract}

\section{Introduction}




In recent years, data-driven methods, and deep learning in particular, have achieved remarkable success across a wide range of learning tasks and data modalities. Among them, sequence modeling occupies a distinctive position at the intersection of time series analysis, natural language processing, dynamical systems, and machine learning \citep{Elman_1990RNN,zeng2022transformerseffectivetimeseries,llmeffective,brachet2026time}. With the rise of deep learning, language became a major driving force behind advances in sequence modeling. Recurrent neural networks (RNNs) \citep{Elman_1990RNN,Jordan_1997b}, Long-short term memory (LSTM) \citep{lstm}, gated architectures \citep{chung2014empiricalevaluationgatedrecurrent}, attention mechanisms \citep{bahdanau2016neuralmachinetranslationjointly}, Transformers \citep{vaswani2023attentionneed}, and later State-Space Models (SSMs) \citep{gu2024mambalineartimesequencemodeling} were extensively developed and refined in the context of natural language processing. These architectural advances were subsequently transferred to other sequential domains, including time series modeling \citep{laptev2017time,rnn-tsf-survey2021,zhang2023effectivelymodelingtimeseries,hu2024timessmsimplifyingunifyingstate,jin2024timellmtimeseriesforecasting}.

RNNs were explicitly conceived as dynamical systems, in which an internal state evolves over time as sequential inputs are processed \citep{Elman_1990RNN}. This formulation is closely related to nonlinear state-space representations, which have long been central to control theory and time series analysis \citep{kalman}. Consider a linear time-invariant discrete-time state-space model,
\begin{eqnarray}
\displaystyle \vx_{t+1} = \displaystyle \mA \vx_t + \mB \vu_t, \
\displaystyle \vy_{t+1} = \displaystyle \mC \vx_{t+1} ,
\label{eq:vanilla-kalman-ssm}
\end{eqnarray}
where $t\in\mathbb{R}^+$ is a timestamp, $\displaystyle \vx_t \in \mathbb{R}^{d_x}$ denotes the \textbf{internal state} of the system at time $t$, $\displaystyle \vu_t \in \mathbb{R}^{d_u}$ a \textbf{control input} acting on its dynamics, and $\displaystyle \vy_t  \in \mathbb{R}^{d_y}$ its \textbf{measured output}. $\mA  \in \mathbb{R}^{d_x \times d_x}$, $\mB\in \mathbb{R}^{d_x \times d_u}$, and $\mC \in \mathbb{R}^{d_y \times d_x}$ are, respectively, the state, input, and output matrices. We call the system defined by Equation \ref{eq:vanilla-kalman-ssm} a \textbf{controlled system}. RNNs extend this formulation by replacing the linear state transition with nonlinear and learnable dynamics. More recently, structured SSMs have revisited this formalism to build sequence models that remain recurrent while allowing efficient computation over long sequences \citep{gu2022efficiently,gu2024mambalineartimesequencemodeling}.

This input-state-output interpretation appeared naturally in early recurrent approaches to machine translation, where source words sequentially drove the recurrent state and target words were produced at the output \citep{castano-etal-1997-machine,Castano_Casacuberta_1997,sundermeyer-etal-2014-translation}. From a dynamical-system perspective, the source sequence thus acts as an exogenous input $\vu$, while the translated sequence constitutes the output $\vy$. Later encoder-decoder architectures made this separation explicit by using distinct recurrent networks for source encoding and target generation \citep{rnn-machine-translation,rnn-sutskever}, a natural choice since source and target follow different sequential distributions and need not share the same dynamics.

While this separation is natural in machine translation, Time Series Forecasting (TSF) is fundamentally different: the observed context and the predicted future are two portions of the same process. When recurrent models (RNNs, LSTMs, and SSMs) are directly transferred to TSF, however, the input-output interpretation becomes less natural. 
Unlike machine translation, the input and output are the same processes, shifted in time by one timestamp. The underlying system is no longer controlled, but an \textbf{autonomous system}. However, $\vu$ still plays the role of a control input. Assigning this role to $\vu$ induces a \textbf{modeling mismatch}. Deep autoregressive models \citep{salinas2020deepar} partially solved this issue by mapping inputs in a different representation space than the predicted outputs. However, when observations are no longer available, such models inject the predictions back into the model input: it becomes an autonomous system, whereas it is a controlled one when observations are available, leading to a \textbf{discrepancy at the context-forecasting boundary} that is induced solely by how the dataset is split during learning. A recent SSM-based work, SpaceTime \citep{zhang2023effectivelymodelingtimeseries}, explicitly assumes that $\vu$ and $\vy$ are two different processes and predicts future observations to make predictions. However, based on how SpaceTime predicts $\vu$, the discontinuity remains, and the fact that $\vu$ and $\vy$ are different cannot hold. Recent works thus only use such recurrent networks as encoder \citep{hu2024attractormemorylongtermtime,sst,sde}, but lose at prediction time the ability of such a unit to \textbf{learn how to move forward in time}.

\begin{table}[t]
\centering
\caption{Overview of how different recurrent formulations satisfy the desired consistency properties in the case of time series forecasting. $\checkmark$ (or $\times$) indicates whether the property is satisfied (or not). We propose OSSM, which satisfies all these properties.}
\label{tab:forecasting_reformulations}
\begin{tabular}{lccc}
\toprule
\textbf{Recurrent mode} 
& \textbf{Modeling} 
& \textbf{Time propagation} 
& \textbf{Regime} \\
& \textbf{consistency}
& \textbf{consistency}
& \textbf{consistency} \\
\midrule
Shallow
& $\times$ & $\checkmark$ & $\times$  \\

Deep
& $\checkmark$ & $\checkmark$ & $\times$   \\

Encoder only
& $\checkmark$ & $\times$ & $\times$ \\

SpaceTime~\citep{zhang2023effectivelymodelingtimeseries}
& $\checkmark$ & $\checkmark$ & $\times$ \\

\midrule
\textbf{OSSM (ours)}
& $\boldsymbol{\checkmark}$
& $\boldsymbol{\checkmark}$
& $\boldsymbol{\checkmark}$
 \\
\bottomrule
\end{tabular}
\end{table}

These different designs highlight three desirable consistency properties for recurrent TSF models:
\begin{itemize}
    \item \textbf{Modeling consistency}: the roles assigned to $\vu$ and $\vy$ should be consistent with their interpretation in the forecasting problem;
    \item \textbf{Time propagation consistency}: the learned dynamics should be exploited to propagate the latent state throughout the whole time window;
    \item \textbf{Regime consistency}: the dynamics attributed to the underlying process should remain consistent across the context and forecasting intervals.
\end{itemize}

In this context, we propose a new class of recurrent models called \textbf{Observer State-Space Models} (OSSMs). We take a different perspective: we consider the inputs $\displaystyle \vu$ to be measurements of a dynamical system's outputs $\displaystyle \vy$. This perspective naturally leads us back to state observers from control theory \citep{luenberger1964observing,luenberger1966observers}. 
During the context, available measurements \textbf{correct} the estimated latent trajectory through an observer; once forecasting begins, this correction disappears, while the underlying state transition remains unchanged. Thus, the computation necessarily changes when measurements become unavailable, but the modeled process dynamics are the same in both the context and the horizon interval. This formulation further allows us to leverage tools from control theory to characterize and control the convergence of the estimated latent state. We summarize our contribution in Table \ref{tab:forecasting_reformulations}. We show that simple SSMs and SpaceTime \citep{zhang2023effectivelymodelingtimeseries} can be recovered as a particular instance of the OSSM framework, but present model mismatches.
This work addresses two research questions:

\textbf{RQ1.} Is it possible to define a recurrent model satisfying the three consistency properties? 

\textbf{RQ2.} If so, does such a model allow for better performance on the time series forecasting task?

To address RQ1, we revisit Luenberger observers and standard SSMs, introduce the OSSM formulation, and characterize the relationship between OSSM, conventional SSMs, and SpaceTime. To address RQ2, we conduct controlled comparisons between OSSM and corresponding SSM variants on established short- and long-term forecasting benchmarks, using matched parameter counts and training protocols. Our results show that explicitly separating state propagation from measurement correction can yield substantial forecasting gains without increasing model capacity.

\section{Background and related work}

\paragraph{Observation-driven and free-running dynamics.}
The discrepancy between recurrent dynamics driven by observed values and those obtained in free-running mode due to Teacher Forcing \citep{williams1989learning} has been studied across several sequential modeling domains \citep{bengio_scheduled,professor_forcing}. 
Closely related difficulties arise when reconstructing dynamical systems, particularly in chaotic regimes, where models trained using observed trajectories may rapidly diverge when recursively rolled out without observations \citep{model-free-switch,onthedifficulty2022,herz2026teacher}. 
In TSF, methods avoid or mitigate this transition by adapting learning strategies \citep{teutsch2022flipped}, isolating recurrent dynamics from the forecasting stage \citep{hu2024attractormemorylongtermtime}, refining predictions \citep{k2vae}, or adapting the dynamics once observations are no longer available \citep{zhang2023effectivelymodelingtimeseries}. Our approach differs in that it revisits the role assigned to observations in the state-space formulation itself: we use observations only as measurements that correct the estimated state when available.

\paragraph{Time series as observations of dynamical systems.}
Our work belongs to a growing family of forecasting approaches that views time series as observations of an underlying dynamical process. 
A first line of work builds on Koopman theory \citep{koopman}, which represents the evolution of observables of a nonlinear dynamical system through a linear operator in a generally infinite-dimensional function space \citep{koopa,gupta2024morizwanziglatentspacekoopman,k2vae}. 
Another line of work relies on reconstruction results from nonlinear and chaotic dynamical systems \citep{hu2024attractormemorylongtermtime,deepedm}. 
Other methods introduce dynamical inductive biases through state-space representations \citep{zhang2023effectivelymodelingtimeseries} or explicitly couple temporal evolution with generative mechanisms such as diffusion models \citep{cachay2023dyffusion,guo2025dynamical}.

\paragraph{Learning-based observers.}
Our work is also related to learning-based state observers. KalmanNet and related neural filtering methods learn or adapt measurement-update mechanisms in state-space models with known or partially known dynamics \citep{revach2022kalmannet}, typically in state-estimation settings where a physical model and, during training, latent-state supervision may be available. Neural Luenberger and Kazantzis--Kravaris--Luenberger (KKL) observers \citep{kazantzis1998nonlinear, niazi2022learning} instead focus on reconstructing the state of nonlinear dynamical systems from measurements. These approaches share the principle of correcting a state estimate with observations, but target state estimation rather than output-only time series forecasting. 

Together, these works support a view of time series forecasting in which predicting future observations amounts, at least approximately, to identifying and propagating underlying dynamics from (partial) observations. Our work adopts this perspective and focuses on a complementary question: how should the available observations interact with the estimated dynamical state for forecasting purposes?

\begin{figure}[t]
\centering

\resizebox{0.98\linewidth}{!}{%
\begin{tikzpicture}[
    x=0.72cm,
    y=0.57cm,
    >=Stealth,
    line cap=round,
    line join=round,
    every node/.style={font=\small},
    curve/.style={line width=1.35pt},
    flow/.style={->,line width=1.0pt},
    feedback/.style={->,line width=1.2pt,pred},
]


\def\xL{3.0}
\def\xT{11.0}
\def\xR{19.2}


\draw[gray!45,dashed,line width=.7pt]
    (\xL,6.45) -- (\xR,6.45);

\draw[gray!60,dashed,line width=.8pt]
    (\xT,0.55) -- (\xT,12.75);

\draw[->,line width=.9pt]
    (\xL,0.30) -- (\xR+0.35,0.30);

\node[below=2pt] at (\xT,0.30) {$T$};
\node[anchor=west] at (\xR+0.35,0.30) {$t$};

\node[font=\large]
    at ({(\xL+\xT)/2},-0.25)
    {Context};

\node[font=\large]
    at ({(\xT+\xR)/2},-0.25)
    {Forecast};


\node[font=\Large\bfseries] at (0.80,9.55) {OSSM};
\node[font=\Large\bfseries] at (0.60,5.35) {SSM};

\node[anchor=east,align=center]
    at (2.90,11.05)
    {latent state\\[-2pt]$x$};

\node[anchor=east,align=center]
    at (2.90,8.00)
    {observation\\[-2pt]$y$};

\node[anchor=east,align=center]
    at (2.90,5.05)
    {input\\[-2pt]$u$};

\node[anchor=east,align=center]
    at (2.90,3.50)
    {latent state\\[-2pt]$x$};

\node[anchor=east,align=center]
    at (2.90,1.60)
    {output\\[-1pt]$\hat y$};



\draw[curve,black,dashed]
plot[smooth,tension=.58] coordinates {
    (3.30,11.25)
    (4.10,11.78)
    (4.90,11.35)
    (5.65,11.12)
    (6.35,11.72)
    (7.05,11.25)
    (7.70,11.55)
    (8.45,11.05)
    (9.20,11.70)
    (10.00,11.35)
    (10.75,10.98)
    (11.50,11.15)
    (12.35,11.55)
    (13.15,11.30)
    (14.00,11.82)
    (14.80,12.25)
    (15.55,11.75)
    (16.35,11.92)
    (17.20,12.55)
    (18.05,12.15)
    (18.90,12.65)
};

\draw[curve,pred]
plot[smooth,tension=.58] coordinates {
    (3.30,10.75)
    (4.10,11.52)
    (4.90,11.18)
    (5.65,11.00)
    (6.35,11.58)
    (7.05,11.18)
    (7.70,11.48)
    (8.45,11.00)
    (9.20,11.58)
    (10.00,11.27)
    (10.75,10.96)
    (11.50,11.12)
    (12.35,11.48)
    (13.15,11.23)
    (14.00,11.70)
    (14.80,12.13)
    (15.55,11.66)
    (16.35,11.84)
    (17.20,12.42)
    (18.05,12.05)
    (18.90,12.52)
};

\node[anchor=west] at (18.98,12.80) {$x_t$};
\node[anchor=west,text=pred] at (18.98,12.40) {$\hat x_t$};


\draw[curve,obs]
plot[smooth,tension=.55] coordinates {
    (3.30,8.05)
    (4.15,8.40)
    (5.00,7.86)
    (5.85,8.12)
    (6.65,8.58)
    (7.45,7.78)
    (8.25,8.36)
    (9.05,7.82)
    (9.80,8.18)
    (10.55,7.90)
    (11.00,7.82)
};

\foreach \x/\y in {
    3.30/8.05,
    4.15/8.40,
    5.00/7.86,
    5.85/8.12,
    6.65/8.58,
    7.45/7.78,
    8.25/8.36,
    9.05/7.82,
    9.80/8.18,
    10.55/7.90
}{
    \fill[obs] (\x,\y) circle (1.7pt);
}


\draw[curve,pred,dashed]
plot[smooth,tension=.55] coordinates {
    (3.30,7.56)
    (4.15,7.90)
    (5.00,7.46)
    (5.85,7.66)
    (6.65,8.06)
    (7.45,7.34)
    (8.25,7.90)
    (9.05,7.38)
    (9.80,7.78)
    (10.55,7.46)
    (11.00,7.44)
};

\foreach \x/\ya/\yb in {
    4.15/8.40/7.90,
    5.00/7.86/7.46,
    5.85/8.12/7.66,
    6.65/8.58/8.06,
    7.45/7.78/7.34,
    8.25/8.36/7.90,
    9.05/7.82/7.38,
    9.80/8.18/7.78
}{
    \draw[<->,gray!65,line width=.65pt]
        (\x,\yb+0.02) -- (\x,\ya-0.02);
}


\foreach \x/\ytop/\ybot in {
    4.15/11.45/7.98,
    5.85/11.10/7.67,
    10.55/11.05/7.52
}{
    \draw[flow,pred]
        (\x,\ytop) -- (\x,\ybot);
}


\draw[flow,obs]
    (4.15,8.42)
    to[out=84,in=-100]
    (4.48,11.37);

\draw[flow,obs]
    (5.85,8.15)
    to[out=84,in=-100] node[pos=.56,left=2pt,fill=white,inner sep=1.5pt,
    text=obs,font=\scriptsize] {correction}
    (6.18,11.38);

\draw[flow,obs]
    (10.55,7.94)
    to[out=82,in=-100]
    (10.82,11.00);


\draw[curve,obs,dashed]
plot[smooth,tension=.55] coordinates {
    (11.00,7.82)
    (11.80,7.48)
    (12.60,7.82)
    (13.40,8.43)
    (14.20,7.92)
    (15.00,7.42)
    (15.80,8.25)
    (16.60,7.78)
    (17.40,8.42)
    (18.20,8.05)
    (18.90,8.62)
};

\draw[curve,pred,dashed]
plot[smooth,tension=.55] coordinates {
    (11.00,7.92)
    (11.80,7.61)
    (12.60,7.96)
    (13.40,8.54)
    (14.20,8.02)
    (15.00,7.55)
    (15.80,8.36)
    (16.60,7.90)
    (17.40,8.52)
    (18.20,8.16)
    (18.90,8.72)
};

\foreach \x/\ytop/\ybot in {
    11.70/11.10/7.65,
    14.00/11.66/8.12,
    17.50/12.30/8.52,
    18.50/12.40/8.38
}{
    \draw[flow,pred]
        (\x,\ytop) -- (\x,\ybot);
}

\node[anchor=west,text=pred]
    at (18.88,8.70)
    {$\hat y_t$};



\draw[curve,obs]
plot[smooth,tension=.55] coordinates {
    (3.30,5.00)
    (4.15,5.35)
    (5.00,4.81)
    (5.85,5.07)
    (6.65,5.53)
    (7.45,4.73)
    (8.25,5.31)
    (9.05,4.77)
    (9.80,5.13)
    (10.55,4.85)
};

\foreach \x/\y in {
    3.30/5.00,
    4.15/5.35,
    5.00/4.81,
    5.85/5.07,
    6.65/5.53,
    7.45/4.73,
    8.25/5.31,
    9.05/4.77,
    9.80/5.13,
    10.55/4.85
}{
    \fill[obs] (\x,\y) circle (1.7pt);
}

\draw[red!85!black,line width=2pt]
    (10.85,4.67) -- (11.12,5.00);

\draw[red!85!black,line width=2pt]
    (10.85,5.00) -- (11.12,4.67);

\node[
    text=obs,
    anchor=west,
    font=\small
]
    at (11.32,4.84)
    {$u_t$ unavailable};


\draw[curve,black]
plot[smooth,tension=.55] coordinates {
    (3.30,3.45)
    (4.15,3.75)
    (5.00,3.10)
    (5.85,3.52)
    (6.65,4.00)
    (7.45,3.25)
    (8.25,3.67)
    (9.05,3.15)
    (9.80,3.48)
    (10.55,3.05)
    (11.00,2.98)
    (11.80,3.25)
    (12.60,3.83)
    (13.40,3.62)
    (14.20,3.15)
    (15.00,2.90)
    (15.80,3.55)
    (16.60,3.10)
    (17.40,3.58)
    (18.20,3.25)
    (18.90,3.82)
};

\node[anchor=west]
    at (18.98,3.83)
    {$x_t$};


\draw[curve,pred,dashed]
plot[smooth,tension=.55] coordinates {
    (3.30,1.55)
    (4.15,1.87)
    (5.00,1.35)
    (5.85,1.66)
    (6.65,2.12)
    (7.45,1.38)
    (8.25,1.84)
    (9.05,1.32)
    (9.80,1.65)
    (10.55,1.35)
    (11.00,1.30)
    (11.80,1.62)
    (12.60,2.17)
    (13.40,1.96)
    (14.20,1.45)
    (15.00,1.20)
    (15.80,1.88)
    (16.60,1.42)
    (17.40,1.94)
    (18.20,1.58)
    (18.90,2.10)
};

\node[anchor=west,text=pred]
    at (18.88,2.08)
    {$\hat y_t$};


\foreach \x/\yu/\yx in {
    4.15/5.30/3.78,
    5.85/5.02/3.58,
    6.65/5.47/4.03,
    8.25/5.26/3.72,
    9.80/5.08/3.52
}{
    \draw[flow,obs]
        (\x,\yu) -- (\x,\yx);
}

\foreach \x/\yx/\yy in {
    4.15/3.70/1.90,
    5.85/3.50/1.72,
    6.65/3.95/2.16,
    8.25/3.63/1.90,
    9.80/3.44/1.70
}{
    \draw[flow,black]
        (\x,\yx) -- (\x,\yy);
}


\foreach \x/\yx/\yy in {
    11.70/3.20/1.68,
    14.40/3.08/1.42,
    16.20/3.45/1.78,
    18.10/3.22/1.60
}{
    \draw[flow,pred]
        (\x,\yx) -- (\x,\yy);
}


\draw[feedback]
    (11.70,1.66)
    to[out=75,in=-100]
    (12.00,3.42);

\draw[feedback]
    (14.40,1.42)
    to[out=74,in=-100]
    (14.72,3.20);

\draw[feedback]
    (16.20,1.80)
    to[out=76,in=-100]
    (16.52,3.30);

\draw[feedback]
    (18.10,1.60)
    to[out=76,in=-100]
    (18.42,3.56);

\end{tikzpicture}%
}

\caption{
Schematic information flow in OSSM and SSM forecasting.
In OSSM, observations correct an autonomous latent-state estimate during
context; after $T$, the correction is removed while the latent dynamics remain
unchanged. In SSM, observations act as inputs during context; after $T$, the
input is unavailable and predicted outputs are fed back into the dynamics.
}
\label{fig:ossm_ssm}
\end{figure}
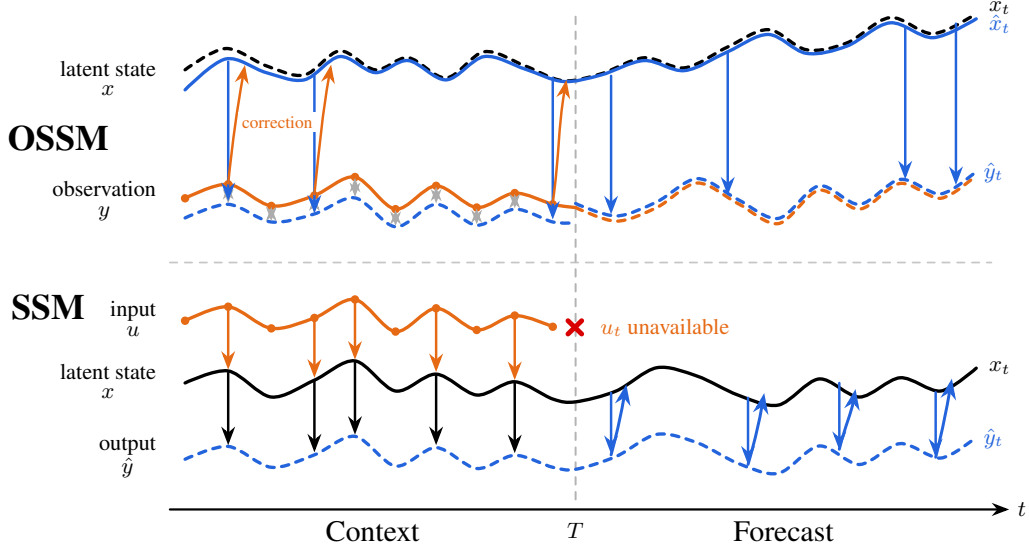

\section{Observer State-Space Models}
\label{sec:ossm}

In this section, we introduce \textbf{Observer State-Space Models (OSSMs)}, a class of models that aims to provide a recurrent deep learning model that validates the three consistency properties for the time series forecasting task.

\subsection{A consistent recurrent model}

Let $\vy$ denote a process from which we have $T\in \mathbb{N}$ historical records $\vy_{1:T}$. In the time series forecasting task, based on these historical records, we aim to predict the $H\in\mathbb{N}$ future ones. To perform the forecasts, we formulate hypotheses to define our deep learning model. 

\begin{hypothesis} \label{hyp1}
    The process $\vy$ is considered to be derived from an underlying dynamical system. 
\end{hypothesis}

Following works based on Koopman theory \citep{koopman}, we formulate the following hypothesis:

\begin{hypothesis}
\label{hyp:3.2}
    There exists a latent space where the input process is derived from an underlying linear dynamical system. 
\end{hypothesis}

Let $\mathcal{E}:\mathbb{R}^{d_y}\to\mathbb{R}^{d_h}$ be a learnable non linear encoder, with $d_y\in\mathbb{N}$ and $d_h\in\mathbb{N}$ the input and hidden dimensions. We thus suppose that $\mathcal{E}_\theta(\vy)$ is derived from a linear latent dynamical system $\vx$ as it follows: 
\begin{eqnarray}
    \vx_{t+1}= \mA_O\vx_t, \quad \mathcal{E}(\vy_t)=\mC_O\vx_t
\end{eqnarray}

with $\mA_O \in \mathbb{R}^{d_x\times d_x}$ the latent dynamical matrix, $d_x\in\mathbb{N}$ the latent dynamical state dimension, and $\mC_O\in\mathbb{R}^{d_h \times d_x}$ the latent measurement matrix.

During the learning procedure, the encoder and the latent matrices weights $(\theta_\mathcal{E},\theta_{A_O},\theta_{C_O})$ are learned. However, learning the latent dynamics is not sufficient. Indeed, the initial conditions of the latent system's state are unknown. Following the Luenberger observer theory \citep{luenberger1964observing,luenberger1966observers}, we use the available observations to correct the latent state estimation in order to recover the true latent trajectory:

\begin{eqnarray}\label{eq:ossm-eq}
    \hat{\vx}_{t+1}= \mA_O\hat{\vx}_t + m_t\mL_O(\mathcal{E}(\vy_t)-\mathcal{E}(\hat{\vy}_t)), \quad \mathcal{E}(\hat{\vy}_t)=\mC_O\hat{\vx}_t
\end{eqnarray}

with $\mL_O \in \mathbb{R}^{d_x\times d_h}$ the latent observer matrix, and $m_t\in \{0,1\}$, a measurement coefficient equal to $1$ when a measurement is available, here $\forall t\leq T$, and $0$ otherwise. Equation \ref{eq:ossm-eq} defines what we call the Observer State-Space Model. As $\mathcal{E}(\hat{\vy})$ lives in the latent space, one should define a nonlinear decoder $\mathcal{D}:\mathbb{R}^{d_h}\to\mathbb{R}^{d_y}$ so the predictions made by the OSSM are projected back in the original space observation. 

However, in classical recurrent formulations, the prediction head both propagates the latent state and performs the predictions in the original space. Equation \ref{eq:ossm-eq} only accounts for the propagation. An OSSM for both is defined as follows:

\begin{eqnarray}
\label{eq:ossm-eq-pred}
     \hat{\vx}_{t+1}= \mA_O\hat{\vx}_t + m_t\mL_O(\mathcal{E}(\vy_t)-\hat{\vy}_t), \quad \hat{\vy}_t=\mC_O\hat{\vx}_t
\end{eqnarray}
Equation~\ref{eq:ossm-eq-pred} adopts the prediction-head parameterization used in
our forecasting architecture. The readout $\hat{\vy}_t=\mC_O\hat{\vx}_t$ is
used both as the prediction produced by the head and as the reference term in
the innovation. Consequently, we set $d_y=d_h$, so that
$\mathcal{E}(\vy_t)-\hat{\vy}_t$ is well-defined. In this parameterization,
$\mathcal{E}$ should be understood as a learned preprocessing of the observed
signal within the shared prediction-head coordinate space, rather than as a
separate decoder--encoder pair. The learnable parameters are
$(\theta_{\mathcal E}, \theta_{\mA_O}, \theta_{\mC_O}, \theta_{\mL_O})$.

\paragraph{Consistency properties}

\begin{proposition}[Modeling consistency]
    OSSM satisfies the modeling consistency property.
\end{proposition}
In the OSSM formalism, the correction term
$(\mathcal{E}(\vy_t)-\hat{\vy}_t)$ compares an observed measurement
representation with its estimate in the same space. The observed and predicted
processes are modeled as deriving from the same latent dynamical system. Their
roles are therefore consistent with their theoretical interpretation. OSSM is
thus consistent in terms of modeling.

\begin{proposition}[Time propagation consistency]
    OSSM satisfies the time propagation consistency property.
\end{proposition}
OSSM moves forward in time according to the latent dynamical matrix $\mA_O$ on
both the context and forecast intervals. The prediction head, therefore, applies
the same time-propagation mechanism to historical and predicted data. OSSM is
thus consistent in terms of time propagation.

\begin{proposition}[Regime consistency]
    OSSM satisfies the regime consistency property.
\end{proposition}
OSSM uses the same latent dynamical matrix $\mA_O$ on both the context and
forecast intervals, whether observations are available or not. The transition
to the forecast regime only removes the measurement-correction term; it does
not introduce different latent dynamics. OSSM is thus consistent across
regimes.

Satisfying the three consistency properties, OSSM provides an affirmative
answer to RQ1.

\paragraph{Luenberger observer properties}
OSSM, in its formulation in Equation \ref{eq:ossm-eq-pred}, is inspired by the Luenberger observer formalism.\footnote{Formulation in Equation \ref{eq:ossm-eq} is closer to the Luenberger observer formalism as the residual term is the comparison of two latent observations, while in Equation \ref{eq:ossm-eq-pred}, a latent observation and one in the original space are compared.} 
Under Hypothesis \ref{hyp:3.2} and while measurements are available, the observer formulation characterizes the evolution of the state-estimation error.
Let $\ve_t=\vx_t-\hat{\vx}_t$ denote the error term between the true latent dynamics and the estimated ones. We have the following recurrence:

\begin{align}
    \ve_{t+1}
    &= \vx_{t+1}-\hat{\vx}_{t+1} \nonumber \\
    &= \mA_O\vx_t
    - \left[
        \mA_O\hat{\vx}_t
        + m_t\mL_O\bigl(\mathcal{E}(\vy_t)-\hat{\vy}_t\bigr)
    \right] \nonumber \\
    &= \mA_O(\vx_t-\hat{\vx}_t)
    - m_t\mL_O\mC_O(\vx_t-\hat{\vx}_t) \nonumber \\
    &= \left(\mA_O-m_t\mL_O\mC_O\right)\ve_t.
    \label{eq:error_dynamics}
\end{align}
During the observed context, $m_t=1$, so
\[
\ve_{t+1}=(\mA_O-\mL_O\mC_O)\ve_t.
\]
Consequently, if the spectral radius $\rho$ of $(\mA_O-\mL_O\mC_O)$ is lower than one, the error term can converge to zero, so the estimated system state can converge to the true one. In the Luenberger observer theory, the matrices $\mA_O$ and $\mC_O$ are already known. If $(\mA_O,\mC_O)$ is observable (see Appendix \ref{app:luenberger} for further details), then one can freely set the weights of $\mL_O$ so the speed of convergence of the estimated state can be controlled.

\paragraph{Learning both to propagate and to correct}
In contrast to classical observer design, the latent matrices that define the system are not known beforehand. OSSMs jointly learn the latent dynamics $\mA_O$, the measurement map $\mC_O$, and the observer gain $\mL_O$. Observability does not need to be explicitly enforced: observable pairs $(\mA_O,\mC_O)$ form an open dense subset of the parameter space and are therefore obtained almost surely under continuous random initialization. In the experiments, $(\mA_O,\mC_O)$ are initialized and $\mL_O$ is set so the model starts with a chosen $\rho(\mA_O-\mL_O\mC_O)$.


\subsection{Coupled Dynamics in SSMs}
\label{sec:model_mismatch}

\paragraph{SSM formulation}

A linear State-Space Model (SSM) defines the recurrent transformation
\begin{align}
    \vx_{t+1} &= \mA_S\vx_t + \mB_S\vu_t, \label{eq:ssm_state}\\
    \vy_{t+1} &= \mC_S\vx_{t+1},          \label{eq:ssm_output}
\end{align}
with $\vu\in\mathbb{R}^{d_u}$ the input control process and $\mB\in\mathbb{R}^{d_x\times d_u}$ is the control matrix. When observations are not available anymore, $\vy$ is fed to the system state equation through $\mB$. During the learning process, $\vy$ is learned to reproduce $\vu$.

\paragraph{SSM as an entangled OSSM}

A closed-loop SSM can be expressed exactly as an OSSM under a change of coordinates. This equivalence is central to our formulation: OSSM does not enlarge the class of linear recurrent prediction heads, but directly parameterizes the autonomous dynamics used during forecasting separately from measurement correction.

Indeed, by setting 

\begin{align}
  \displaystyle \mA_O &= \displaystyle \mA_S + \displaystyle \mB_S\displaystyle \mC_S \\
  \displaystyle \mL_O &= \displaystyle \mB_S \\
  \displaystyle \mC_O &= \displaystyle \mC_S
\end{align}
We can recover the SSM as an OSSM. Similarly, SpaceTime \citep{zhang2023effectivelymodelingtimeseries}, an SSM designed specifically for time series forecasting purposes, can be rewritten as an OSSM (see Appendix \ref{app:spacetime}).\footnote{A recent analysis also connected SpaceTime to Luenberger observers under adversarial attacks \citep{anand2026adversarialrobustnessdeepstate}. In contrast, we consider the TSF task and SpaceTime arises as a particular instance of the OSSM class.}

Conventional closed-loop SSMs and OSSMs represent the same class of linear recurrent forecasting heads. Their difference lies in the parameterization: the former represents the context update through an input matrix and readout, whereas OSSM directly parameterizes the autonomous transition used during free-running forecasting and the gain that assimilates available measurements. This separation makes these two roles explicit and independently controllable at initialization.

Rewriting the SSM as an OSSM, while supposing that the observed (encoded) signal $\displaystyle \vu$ is a measurement which corrects the internal state estimation, some modeling mismatches arise:
\begin{itemize}
    \item the evolution matrix $\displaystyle \mA_O$ depends on the observer matrix $\displaystyle \mL_O$. However, $\displaystyle \mA_O$ is independent of the observer, which is exogenous to the studied system and is added to tackle an estimation problem;
    \item the evolution matrix $\displaystyle \mA_O$ also depends on the readout matrix. However, again, the dynamics matrix is also independent of $\displaystyle \mC_O$, which represents the channels that are measured by the user.
\end{itemize}

In classical control, the system dynamics are specified independently of the observer, which is subsequently designed using the dynamics and the available measurement map. Under an observer-based interpretation of conventional SSMs, however, these roles become entangled: the effective autonomous transition depends on parameters that also govern measurement assimilation and readout.

This observation motivates RQ2: Does explicitly separating autonomous propagation from measurement correction improve time series forecasting? We investigate this question through controlled comparisons between conventional SSMs and their observer-parameterized counterparts. Better results from the OSSM model compared to SSM would validate RQ2.

\section{Experiments}

\subsection{Experimental setup}


The OSSM is designed to perform time-series predictions while preserving the dynamics from the context into the prediction interval. In general, SSM architectures consist of SSM encoder units that learn a representation of the input. The final layer reads the encoded input and performs the predictions. 

We compare in the same settings two model families, illustrated in Figure \ref{fig:ossm_ssm}:
\begin{itemize}
    \item an OSSM, which is composed of SSM layers as encoders, and a final OSSM layer as the prediction head.
    \item an SSM, which is composed of the same encoding units, but the prediction head is an SSM.
\end{itemize}

Our experiments address a controlled question: for a fixed recurrent forecasting architecture, optimizer, parameter budget, and training protocol, does the observer parameterization improve the standard SSM prediction head? We evaluate the two model families on the TSF task using the TFB benchmark~\citep{qiu2024tfb}, across four forecasting horizons, considering the ETT datasets, Electricity, Traffic, Solar, Exchange, Weather, and Wind for long-term forecasting, using a lookback window of $512$, and ILI, NYSE, NASDAQ, and NN5 for short-term forecasting, using a lookback window of $104$. Results are aggregated across $6$ seeds and hyperparameter configurations using the median improvement from SSM to OSSM with the same configuration, and the fraction of pairs OSSM won. We report five complementary normalized metrics: MSE, MAE, WAPE, sMAPE, and msMAPE, enabling us to assess the consistency of the results across diverse metrics. Results are reported in Table \ref{tab:ossm_ssm_results}. Detailed and raw results can be found in Appendix \ref{app:raw-results}.

To promote stable free-running dynamics at initialization, we control the spectral radius of the transition used during forecasting: $\rho(\mA_O)$ for OSSM and $\rho(\mA_S+\mB_S\mC_S)$ for the SSM baseline. In both cases, this radius is set to $0.9$.

For OSSM, we additionally initialize the observer-error dynamics $\mA_O-\mL_O\mC_O$ with spectral radius $0.91$. This setting is part of the observer-structured parameterization: it makes both the free-running transition and the measurement-correction dynamics explicit at initialization. All parameters are subsequently optimized without spectral constraints.


\subsection{Results}


\begin{table*}[t]
    \centering
    \caption{Paired OSSM-SSM comparison over 5,376 runs (8 configurations, 4 horizons, and 6 seeds per dataset and head). We report the relative gain
$\Delta_M=100(M_{\mathrm{SSM}}-M_{\mathrm{OSSM}})/M_{\mathrm{SSM}}$
as mean $\pm$ s.d. across seeds. Blue/red bold indicates a strictly positive/negative $1\sigma$ interval; black indicates overlap with zero. Win rate is pooled across all five metrics; a diverged head loses all five. Relative gains use only pairs where both heads converge, and thus conservatively underestimate OSSM's advantage.}
    \label{tab:ossm_ssm_results}
    \small
    \setlength{\tabcolsep}{3.6pt}

    \begin{tabular}{lcccccc}
        \toprule
        \textbf{Dataset}
        & \textbf{MSE}
        & \textbf{MAE}
        & \textbf{WAPE}
        & \textbf{sMAPE}
        & \textbf{msMAPE}
        & \textbf{Win rate} \\
        \midrule

        \multicolumn{7}{l}{\textit{Short-term forecasting}} \\
        \addlinespace[2pt]

        NN5
        & \result{16.1}{2.1}
        & \result{11.7}{1.7}
        & \result{11.9}{1.7}
        & \result{13.5}{2.0}
        & \result{13.6}{2.0}
        & 94.9\% \\

        NYSE
        & \result{9.4}{8.4}
        & \result{8.5}{4.3}
        & \result{7.1}{2.8}
        & \result{6.8}{3.2}
        & \result{7.3}{3.3}
        & 71.0\% \\

        ILI
        & \result{3.1}{5.7}
        & \result{2.9}{3.8}
        & \result{3.4}{3.8}
        & \result{3.9}{4.7}
        & \result{4.0}{5.4}
        & 68.6\% \\

        NASDAQ
        & \result{-1.7}{6.0}
        & \result{1.1}{3.6}
        & \result{1.2}{3.0}
        & \result{0.6}{2.4}
        & \result{0.6}{2.5}
        & 49.0\% \\

        \midrule
        \multicolumn{7}{l}{\textit{Long-term forecasting}} \\
        \addlinespace[2pt]

        Electricity
        & \result{22.9}{4.6}
        & \result{17.1}{3.1}
        & \result{17.0}{3.1}
        & \result{17.6}{3.1}
        & \result{18.4}{3.3}
        & 89.9\% \\

        Solar
        & \result{15.0}{2.8}
        & \result{8.9}{1.7}
        & \result{8.9}{1.7}
        & \result{10.0}{2.1}
        & \result{10.2}{2.1}
        & 69.1\% \\

        Traffic
        & \result{12.2}{2.5}
        & \result{11.8}{3.0}
        & \result{11.8}{2.9}
        & \result{14.9}{4.2}
        & \result{14.8}{4.1}
        & 88.0\% \\

        ETTh1
        & \result{7.2}{1.6}
        & \result{3.9}{0.7}
        & \result{3.9}{0.8}
        & \result{4.2}{1.0}
        & \result{5.1}{1.1}
        & 66.4\% \\

        ETTm1
        & \result{3.8}{4.8}
        & \result{2.0}{2.1}
        & \result{2.1}{2.0}
        & \result{1.8}{1.7}
        & \result{2.3}{2.0}
        & 59.5\% \\

        Wind
        & \result{2.6}{1.5}
        & \result{1.7}{0.9}
        & \result{1.8}{1.0}
        & \result{1.3}{0.6}
        & \result{1.4}{0.6}
        & 59.0\% \\

        ETTm2
        & \result{1.3}{0.8}
        & \result{0.7}{0.3}
        & \result{0.6}{0.3}
        & \result{0.9}{0.2}
        & \result{1.0}{0.2}
        & 68.8\% \\

        Exchange
        & \result{0.5}{5.8}
        & \result{2.2}{2.3}
        & \result{3.9}{2.3}
        & \result{2.1}{1.2}
        & \result{3.1}{1.4}
        & 67.9\% \\

        ETTh2
        & \result{-0.6}{1.5}
        & \result{-0.2}{0.3}
        & \result{-0.2}{0.4}
        & \result{0.5}{0.4}
        & \result{0.3}{0.4}
        & 57.9\% \\

        Weather
        & \result{-1.6}{9.1}
        & \result{1.3}{0.9}
        & \result{1.3}{0.8}
        & \result{0.9}{0.5}
        & \result{1.2}{0.7}
        & 54.3\% \\

        \bottomrule
    \end{tabular}
\end{table*}

\paragraph{Controlled comparison.}
Table~\ref{tab:ossm_ssm_results} compares OSSM with its matched SSM counterpart across the full experimental grid. The two models have identical encoders, parameter counts, optimization settings, data splits, and random seeds; they differ only in the parameterization of the forecasting head. Consequently, the comparison isolates whether separating autonomous propagation from measurement correction is beneficial.

OSSM improves the five evaluation metrics on most datasets, with the largest and most consistent gains on Electricity, Traffic, NN5, Solar, and ETTh1. These improvements are especially pronounced at longer horizons, where errors accumulated during free-running prediction are more consequential. The effect is not universal: NASDAQ, ETTh2, and Weather exhibit near-parity or mixed outcomes. This heterogeneity is expected, since the benefit of observer correction depends on how informative the available context is about the latent state.

\paragraph{Robustness.}
Besides average error, OSSM is more reliable during optimization. Over the 5,376 matched runs, 96 OSSM runs diverged, compared with 225 SSM runs. Since both heads have the same capacity and training protocol, this difference suggests that the observer parameterization yields a more favorable recurrent optimization geometry and improves average forecasting accuracy across many datasets.

\paragraph{Interpretation.}
The results support our second research question in a controlled setting: explicitly distinguishing the dynamics used to propagate the state from the mechanism that assimilates observed measurements can improve recurrent forecasting. They do not imply that OSSM is uniformly preferable for every dataset, nor do they constitute a comparison with all modern TSF architectures.

\section{Conclusion and future work}
Observer State-Space Models (OSSMs) provide a recurrent formulation for time series forecasting in which observations serve as measurements that correct an estimated latent state, rather than as control inputs that define its dynamics. A single autonomous transition is retained across the observed context and the free-running forecast horizon, thereby satisfying the modeling, time-propagation, and regime-consistency properties introduced in this paper. This construction provides a constructive answer to RQ1. It also offers a unified interpretation of conventional SSMs and SpaceTime, revealing that their forecasting updates entangle latent-state propagation with measurement-related terms.

RQ2 is addressed through a strictly controlled evaluation across 14 datasets, 4 horizons, 8 configurations, and 6 seeds. The OSSM and SSM models share the same encoders, parameter counts, optimization protocol, data splits, random seeds, and initialization of free-running stability; only the forecasting-head parameterization differs. Across this 5,376-run grid, OSSM improves performance on most datasets and across five complementary metrics, with particularly consistent gains on Electricity, Traffic, NN5, Solar, and ETTh1. It also substantially reduces optimization failures, with 96 divergent runs versus 225 for the matched SSM baseline.

The results do not establish universal superiority over all forecasting architectures or datasets. Instead, they provide controlled evidence that explicitly separating state propagation from measurement correction is a useful inductive bias for recurrent forecasting, especially when the observed context is informative about the latent state. Beyond contiguous forecasting, OSSMs naturally extend to intermittent observations, where autonomous propagation bridges missing intervals and new measurements correct the state estimate. Future work can combine this observer structure with stronger encoders and a more general SpaceTime-compatible formulation.

\subsection*{AI use statement}

In preparing this manuscript, we occasionally used suggestions from LLMs to improve clarity, grammar, and overall readability in both the main content and the code provided. We also used it to search for papers in the related work, especially on chaotic system identification and early adaptation of recurrent models in time series forecasting. All scientific content, including experimental design, data analysis, results, and interpretations, is independently developed by the authors.

\subsection*{Ethics statement}

This work is foundational research on already existing open-source models, tested on a public benchmark. This work has no impact on society at large, beyond aiming to get a better understanding of deep learning models. This work could reduce model energy consumption by guiding research on dynamic modeling.

\subsection*{Reproducibility statement}

For transparency and reproducibility, as detailed in the supplementary material in Appendix \ref{app:code-implementation-details}, we base our code on the public repository developed by the TFB benchmark. Along with the paper, we provide the full code, including the studied models and their hyperparameters, used to evaluate them on the TSF task; it is included in the supplementary material.



\bibliography{my_refs}
\bibliographystyle{iclr2027_conference}

\newpage

\appendix

\section{Luenberger theoretical details} \label{app:luenberger}

The convergence of the state estimate is governed by the eigenvalues of $\mA-\mL\mC$. The pair $(\mA,\mC)$ is observable if
\begin{equation}
    \operatorname{rank}
    \begin{bmatrix}
        \mC^\top &
        (\mC\mA)^\top &
        \cdots &
        (\mC\mA^{d_x-1})^\top
    \end{bmatrix}^{\!\top}
    = d_x.
\end{equation}
When this condition holds, the eigenvalues of $\mA-\mL\mC$ can be freely set through $\mL$, so the estimation error convergence speed can be controlled. 

\section{SpaceTime through the OSSM formulation}
\label{app:spacetime}

\subsection{SpaceTime forecasting model}

SpaceTime \citep{zhang2023effectivelymodelingtimeseries} introduces an SSM architecture specifically designed for time series modeling. A central difference with standard deep SSMs is the parameterization of the transition matrix as a companion matrix. This structure provides a canonical representation of discrete-time autoregressive processes and allows the resulting SSM to represent AR dynamics that standard SSM parameterizations may fail to capture.

For forecasting, SpaceTime further introduces a closed-loop decoder. We focus here on this mechanism, as it is the component relevant to our OSSM interpretation. They suppose that the input and output are derived from two distinct processes. Thus, when measurements are no longer available, they don't feed the model with their outputs but with an estimate of the input process. Let $\bar{\vu}_t$ denote the representation received by the decoder. Its state evolves as
\begin{align}
\vx_{t+1}
= \mA_{\mathrm{ST}}\vx_t
+ \mB_{\mathrm{ST}}\bar{\vu}_t, \quad
\hat{\vy}_{t+1}
= \mC_{\mathrm{ST}}\vx_{t+1}
\end{align}
In addition, SpaceTime learns a projection $\mK_{\mathrm{ST}}$ to reconstruct the next decoder input,
\begin{align}
\hat{\bar{\vu}}_{t+1}
= \mK_{\mathrm{ST}}\vx_{t+1}
\end{align}

The resulting autonomous dynamics are therefore
\begin{align}
\vx_{t+1}
=
\left(
\mA_{\mathrm{ST}}
+ \mB_{\mathrm{ST}}\mK_{\mathrm{ST}}
\right)\vx_t
\end{align}

\subsection{OSSM reformulation}

Following the computations performed in Section \ref{sec:model_mismatch}, SpaceTime can be written as a particular representation-space OSSM. Since $\bar{\vu}_t$ is the decoder input representation, while $\vy_t$ is the output process, we distinguish the map used in the innovation term from the output readout:
\begin{align}
    \mA_O &= \mA_{\mathrm{ST}} + \mB_{\mathrm{ST}}\mK_{\mathrm{ST}}, \\
    \mL_O &= \mB_{\mathrm{ST}}, \\
    \mK_O &= \mK_{\mathrm{ST}}, \\
    \mC_O &= \mC_{\mathrm{ST}}.
\end{align}
Here, $\mK_O$ reconstructs the decoder input representation $\bar{\vu}_t$, whereas $\mC_O$ remains the output readout, consistently with the OSSM formulation.

Indeed, in the context,
\begin{align}
    \hat{\vx}_{t+1} &= \left(\mA_{\mathrm{ST}} + \mB_{\mathrm{ST}}\mK_{\mathrm{ST}}\right)\hat{\vx}_t + \mB_{\mathrm{ST}}\left(\bar{\vu}_t - \mK_{\mathrm{ST}}\hat{\vx}_t\right) \\
    &= \mA_{\mathrm{ST}}\hat{\vx}_t + \mB_{\mathrm{ST}}\bar{\vu}_t,
\end{align}
which is the original SpaceTime update. The corresponding output is $\hat{\vy}_{t+1}=\mC_{\mathrm{ST}}\hat{\vx}_{t+1}$. Once observations are removed, $m_t=0$ and the OSSM evolves autonomously as
\begin{align}
    \hat{\vx}_{t+1} = \left(\mA_{\mathrm{ST}} + \mB_{\mathrm{ST}}\mK_{\mathrm{ST}}\right)\hat{\vx}_t,
\end{align}
which is precisely the closed-loop transition used by SpaceTime for forecasting.

This reformulation exposes the same limitation as in the SSM formulation: the autonomous forecasting transition is coupled to the innovation parameterization through $\mB_{\mathrm{ST}}\mK_{\mathrm{ST}}$. Hence, the free-running dynamics are not learned independently of the map used to predict the decoder input representation.

Moreover, although $\mA_{\mathrm{ST}}$ is constrained to a companion form, the transition governing free-running forecasting, $\mA_{\mathrm{ST}}+\mB_{\mathrm{ST}}\mK_{\mathrm{ST}}$, is generally not. The companion constraint therefore characterizes the context update, whereas the autonomous dynamics used in practice during forecasting follow a different, unconstrained transition.

This correspondence also highlights a limitation of SpaceTime as an OSSM instance: the innovation is defined in the decoder-input representation space through $\bar{\vu}_t-\mK_O\hat{\vx}_t$, whereas the final prediction is read out through $\mC_O\hat{\vx}_t$. A more general OSSM formulation may therefore parameterize the innovation map and the output readout independently. Evaluating this decoupled parameterization within the complete SpaceTime architecture would require a dedicated parameter-matched experimental study, which we leave for future work.

\section{Code and implementation Details}\label{app:code-implementation-details}

\paragraph{Code} The code is available in the supplementary. The code is based on the repository\footnote{https://github.com/decisionintelligence/TFB (pulled in September 2026)} developed for the TFB benchmark~\citep{qiu2024tfb} under the MIT license. We only provide the \path{\ossm} folder. To run the model, one should pull the TFB repository and add the ossm folder into the \path{\ts_benchmark\baselines\} folder. Models are trained using the standard \path{deep_forecasting_model_base.py} file. In the \path{\ossm\functional\} folder, we implement the functions developed in the SpaceTime \citep{zhang2023effectivelymodelingtimeseries} repository.\footnote{https://github.com/HazyResearch/spacetime/}

Everything was done in accordance with the MIT license, which grants the right to use the code and datasets of the TFB benchmark without restriction.

Everything was done under the Apache-2.0 license for the use of the SpaceTime code.

\paragraph{Resources and implementation} All the considered models were trained on Rocky Linux 8.10, on either:
\begin{itemize}
    \item Intel Xeon Gold 6230 20C @ 2.1GHz with 768 GB memory with up to four NVIDIA Tesla V100 with 32 GB VRAM,
    \item Intel Xeon Gold 6346 16C @ 3.1GHz with 1024 GB memory with up to four Nvidia HGX A100 with 40 GB memory.
\end{itemize}

Training was done in PyTorch\footnote{Torch version was $2.5.1$ with cuda toolkit $12.1$ and Python $3.11$}, using the default learning procedure set in the \path{deep_forecasting_model_base.py} file. Hyperparameters can be found in the \path{.\scripts} folder presented in \texttt{.sh} files and detailed model configuration in Appendix \ref{app:raw-results}. Datasets are split following the rolling forecast method to prevent information leakage from the future to the past. We needed at most $1$ GPUs with $15$ CPUs per task, running for several minutes.

\section{Experiment details}

\subsection{Details on the Datasets} \label{app:datasets}

We provide, in Table \ref{tab:datasets}, the detailed statistics of the datasets used in our experiments. It includes the domain, the sampling frequency, the variate dimension, and the data split for training, validation, and testing.

\begin{table}[h]
\caption{Statistics of multivariate datasets of the TFB benchmark~\citep{qiu2024tfb} taken from~\citep{qiu2025duetdualclusteringenhanced}.}
\resizebox{0.999\linewidth}{!}{
\centering
\begin{tabular}{lllllll}
\hline
\textbf{Dataset} & \textbf{Domain} & \textbf{Frequency} & \textbf{Lengths} & \textbf{Dim} & \textbf{Split} & \textbf{Description} \\ \hline
Traffic & Traffic & 1 hour & 17,544 & 862 & 7:1:2 & Road occupancy rates measured by 862 sensors on San Francisco Bay area freeways \\ 
ETTh1 & Electricity & 1 hour & 14,400 & 7 & 6:2:2 & Power transformer 1, comprising seven indicators such as oil temperature and useful load \\ 
ETTh2 & Electricity & 1 hour & 14,400 & 7 & 6:2:2 & Power transformer 2, comprising seven indicators such as oil temperature and useful load \\ 
ETTm1 & Electricity & 15 mins & 57,600 & 7 & 6:2:2 & Power transformer 1, comprising seven indicators such as oil temperature and useful load \\ 
ETTm2 & Electricity & 15 mins & 57,600 & 7 & 6:2:2 & Power transformer 2, comprising seven indicators such as oil temperature and useful load \\ 
Electricity & Electricity & 1 hour & 26,304 & 321 & 7:1:2 & Electricity records the electricity consumption in kWh every 1 hour from 2012 to 2014 \\ 
Solar & Energy & 10 mins & 52,560 & 137 & 6:2:2 & Solar production records collected from 137 PV plants in Alabama \\ 
Wind & Energy & 15 mins & 48,673 & 7 & 7:1:2 & Wind power records from 2020-2021 at 15-minute intervals \\ 
Weather & Environment & 10 mins & 52,696 & 21 & 7:1:2 & Recorded every for the whole year 2020, which contains 21 meteorological indicators \\ 
Exchange & Economic & 1 day & 7,588 & 8 & 7:1:2 & ExchangeRate collects the daily exchange rates of eight countries \\ 
NASDAQ & Stock & 1 day & 1,244 & 5 & 7:1:2 & Records opening price, closing price, trading volume, lowest price, and highest price \\ 
NYSE & Stock & 1 day & 1,243 & 5 & 7:1:2 & Records opening price, closing price, trading volume, lowest price, and highest price \\ 
NN5 & Banking & 1 day & 791 & 111 & 7:1:2 & NN5 is from banking, records the daily cash withdrawals from ATMs in UK \\ 
ILI & Health & 1 week & 966 & 7 & 7:1:2 & Recorded indicators of patients' data from Centers for Disease Control and Prevention \\ 
 \hline
\end{tabular}}
\label{tab:datasets}
\end{table}

\subsection{OSSM vs SSM}

\paragraph{Encoding units}

The encoding units for both models are standard SSM, which are freely learned during the learning procedure:

$$
    \vx_{t+1} = \mA\vx_t + \mB\vu_{t+1}, \quad \vy_{t} = \mC\vx_{t}
$$

A first feed-forward acts on the channel dimension to prepare the data for the SSM units in series. 
Then, each SSM unit maps $\vu^l \to \vu^{l+1}$, where $l$ is the $l^{th}$ encoding layer.

\paragraph{Prediction head} As detailed in the main part, the prediction is either an OSSM or an SSM as described.


%
%
\section{Raw results}
\label{app:raw-results}

We report every individual run behind the aggregate results. The grid is
fully crossed: 14 datasets $\times$ 4 forecast horizons $\times$ 6
random seeds $\times$ 8 architecture configurations $\times$ 2 prediction
heads, for a total of 5376 runs. The observer head (OSSM) and the standard head (SSM)
share the architecture, the optimizer, the data splits and the seeds, so each
(dataset, horizon, seed) cell is a strictly paired comparison, and only the head differs.

All metrics are the normalized MSE and MAE reported by TFB on the test split;
Lower is better. Within each cell, the better of the two heads is in bold.
A run is called \emph{divergent} when its normalized MSE exceeds 10 or when it
produced no metric at all; such runs are kept in the tables, marked
``div.'', and counted as a loss for the head that diverged. Across the whole
grid 321 runs diverged, 225 of them from the SSM head and 96 from
the OSSM head. Where a cell was run more than once, the most recent run is
reported.

\subsection{Configuration 1}

\begin{table}[H]
\centering
\caption{Hyper-parameters of configuration 1. Every other setting follows the
TFB defaults, and the two heads share all of them, including the random seeds.}
\small
\begin{tabular}{lc}
\toprule
Hyper-parameter & Value \\
\midrule
Encoder layers $e_{\mathrm{layers}}$ & 2 \\
State dimension $d_x$ & 64 \\
Model dimension $d_{\mathrm{model}}$ & 64 \\
Learning rate & $10^{-4}$ \\
Batch size & 32 \\
\bottomrule
\end{tabular}
\end{table}

\begingroup
\setlength{\tabcolsep}{2pt}
\footnotesize
\begin{longtable}{l r cc cc cc cc cc cc }
\caption{Normalized MSE for configuration 1, lower is better. Rows are (dataset, forecast horizon $H$) pairs, columns are the six random seeds, each split into the observer head (OSSM) and the standard head (SSM). The better of the two is in bold. ``div.'' marks a run that diverged, i.e.\ reached a normalized MSE above 10 or produced no metric at all; divergence counts as a loss, so the head that converged is shown in bold.}\label{tab:raw-cfg1-mse}\\
\toprule
Dataset & $H$ & \multicolumn{2}{c}{2021} & \multicolumn{2}{c}{2022} & \multicolumn{2}{c}{2023} & \multicolumn{2}{c}{2024} & \multicolumn{2}{c}{2025} & \multicolumn{2}{c}{2026} \\
\cmidrule(lr){3-4}\cmidrule(lr){5-6}\cmidrule(lr){7-8}\cmidrule(lr){9-10}\cmidrule(lr){11-12}\cmidrule(lr){13-14}
 & & OSSM & SSM & OSSM & SSM & OSSM & SSM & OSSM & SSM & OSSM & SSM & OSSM & SSM \\
\midrule
\endfirsthead
\multicolumn{14}{l}{\itshape Table \ref{tab:raw-cfg1-mse}, continued.}\\
\toprule
Dataset & $H$ & \multicolumn{2}{c}{2021} & \multicolumn{2}{c}{2022} & \multicolumn{2}{c}{2023} & \multicolumn{2}{c}{2024} & \multicolumn{2}{c}{2025} & \multicolumn{2}{c}{2026} \\
\cmidrule(lr){3-4}\cmidrule(lr){5-6}\cmidrule(lr){7-8}\cmidrule(lr){9-10}\cmidrule(lr){11-12}\cmidrule(lr){13-14}
 & & OSSM & SSM & OSSM & SSM & OSSM & SSM & OSSM & SSM & OSSM & SSM & OSSM & SSM \\
\midrule
\endhead
\endfoot
\endlastfoot
\multicolumn{14}{l}{\textit{Short-term forecasting}} \\*
\addlinespace[2pt]
NN5 & 24 & \textbf{1.201} & 1.372 & \textbf{1.279} & 1.403 & \textbf{1.257} & 1.378 & \textbf{1.250} & 1.364 & \textbf{1.208} & 1.332 & \textbf{1.205} & 1.354 \\*
 & 36 & \textbf{1.203} & 1.330 & \textbf{1.275} & 1.353 & \textbf{1.230} & 1.336 & \textbf{1.240} & 1.325 & \textbf{1.181} & 1.310 & \textbf{1.175} & 1.319 \\*
 & 48 & \textbf{1.201} & 1.313 & \textbf{1.276} & 1.333 & \textbf{1.218} & 1.320 & \textbf{1.238} & 1.309 & \textbf{1.165} & 1.296 & \textbf{1.157} & 1.307 \\*
 & 60 & \textbf{1.208} & 1.305 & \textbf{1.278} & 1.321 & \textbf{1.212} & 1.312 & \textbf{1.241} & 1.301 & \textbf{1.161} & 1.292 & \textbf{1.150} & 1.302 \\
\midrule
NYSE & 24 & \textbf{0.382} & 1.020 & \textbf{0.274} & 0.631 & 0.424 & \textbf{0.320} & 0.356 & \textbf{0.343} & 0.325 & \textbf{0.287} & 0.392 & \textbf{0.328} \\*
 & 36 & \textbf{1.013} & 1.346 & \textbf{0.424} & 0.714 & \textbf{0.470} & 0.536 & \textbf{1.058} & 1.319 & \textbf{0.563} & 0.565 & 0.553 & \textbf{0.499} \\*
 & 48 & \textbf{0.858} & 1.659 & 1.084 & \textbf{0.996} & \textbf{0.699} & 0.881 & 1.436 & \textbf{1.188} & \textbf{1.549} & 1.638 & 1.541 & \textbf{0.581} \\
\bottomrule
\end{longtable}
\endgroup


\begingroup
\setlength{\tabcolsep}{2pt}
\footnotesize
\begin{longtable}{l r cc cc cc cc cc cc }
\caption{Normalized MAE for configuration 1, lower is better. Rows are (dataset, forecast horizon $H$) pairs, columns are the six random seeds, each split into the observer head (OSSM) and the standard head (SSM). The better of the two is in bold. ``div.'' marks a run that diverged, i.e.\ reached a normalized MSE above 10 or produced no metric at all; divergence counts as a loss, so the head that converged is shown in bold.}\label{tab:raw-cfg1-mae}\\
\toprule
Dataset & $H$ & \multicolumn{2}{c}{2021} & \multicolumn{2}{c}{2022} & \multicolumn{2}{c}{2023} & \multicolumn{2}{c}{2024} & \multicolumn{2}{c}{2025} & \multicolumn{2}{c}{2026} \\
\cmidrule(lr){3-4}\cmidrule(lr){5-6}\cmidrule(lr){7-8}\cmidrule(lr){9-10}\cmidrule(lr){11-12}\cmidrule(lr){13-14}
 & & OSSM & SSM & OSSM & SSM & OSSM & SSM & OSSM & SSM & OSSM & SSM & OSSM & SSM \\
\midrule
\endfirsthead
\multicolumn{14}{l}{\itshape Table \ref{tab:raw-cfg1-mae}, continued.}\\
\toprule
Dataset & $H$ & \multicolumn{2}{c}{2021} & \multicolumn{2}{c}{2022} & \multicolumn{2}{c}{2023} & \multicolumn{2}{c}{2024} & \multicolumn{2}{c}{2025} & \multicolumn{2}{c}{2026} \\
\cmidrule(lr){3-4}\cmidrule(lr){5-6}\cmidrule(lr){7-8}\cmidrule(lr){9-10}\cmidrule(lr){11-12}\cmidrule(lr){13-14}
 & & OSSM & SSM & OSSM & SSM & OSSM & SSM & OSSM & SSM & OSSM & SSM & OSSM & SSM \\
\midrule
\endhead
\endfoot
\endlastfoot
\multicolumn{14}{l}{\textit{Short-term forecasting}} \\*
\addlinespace[2pt]
NN5 & 24 & \textbf{0.854} & 0.944 & \textbf{0.904} & 0.956 & \textbf{0.888} & 0.951 & \textbf{0.883} & 0.943 & \textbf{0.862} & 0.925 & \textbf{0.865} & 0.942 \\*
 & 36 & \textbf{0.867} & 0.934 & \textbf{0.913} & 0.944 & \textbf{0.884} & 0.940 & \textbf{0.888} & 0.934 & \textbf{0.857} & 0.927 & \textbf{0.859} & 0.934 \\*
 & 48 & \textbf{0.871} & 0.932 & \textbf{0.917} & 0.940 & \textbf{0.882} & 0.937 & \textbf{0.892} & 0.931 & \textbf{0.853} & 0.925 & \textbf{0.854} & 0.932 \\*
 & 60 & \textbf{0.878} & 0.931 & \textbf{0.921} & 0.938 & \textbf{0.881} & 0.937 & \textbf{0.897} & 0.931 & \textbf{0.854} & 0.927 & \textbf{0.853} & 0.933 \\
\midrule
NYSE & 24 & \textbf{0.455} & 0.767 & \textbf{0.351} & 0.587 & 0.469 & \textbf{0.419} & 0.432 & \textbf{0.422} & 0.404 & \textbf{0.384} & 0.434 & \textbf{0.422} \\*
 & 36 & \textbf{0.751} & 0.903 & \textbf{0.463} & 0.611 & \textbf{0.485} & 0.558 & \textbf{0.760} & 0.897 & 0.565 & \textbf{0.545} & 0.524 & \textbf{0.516} \\*
 & 48 & \textbf{0.720} & 1.022 & 0.783 & \textbf{0.735} & \textbf{0.605} & 0.730 & 0.922 & \textbf{0.819} & \textbf{0.976} & 1.017 & 0.968 & \textbf{0.559} \\
\bottomrule
\end{longtable}
\endgroup


\subsection{Configuration 2}

\begin{table}[H]
\centering
\caption{Hyper-parameters of configuration 2. Every other setting follows the
TFB defaults, and the two heads share all of them, including the random seeds.}
\small
\begin{tabular}{lc}
\toprule
Hyper-parameter & Value \\
\midrule
Encoder layers $e_{\mathrm{layers}}$ & 2 \\
State dimension $d_x$ & 64 \\
Model dimension $d_{\mathrm{model}}$ & 128 \\
Learning rate & $10^{-4}$ \\
Batch size & 32 \\
\bottomrule
\end{tabular}
\end{table}

\begingroup
\setlength{\tabcolsep}{2pt}
\footnotesize
\begin{longtable}{l r cc cc cc cc cc cc }
\caption{normalized MSE for configuration 2, lower is better. Rows are (dataset, forecast horizon $H$) pairs, columns are the six random seeds, each split into the observer head (OSSM) and the standard head (SSM). The better of the two is in bold. ``div.'' marks a run that diverged, i.e.\ reached a normalized MSE above 10 or produced no metric at all; divergence counts as a loss, so the head that converged is shown in bold.}\label{tab:raw-cfg2-mse}\\
\toprule
Dataset & $H$ & \multicolumn{2}{c}{2021} & \multicolumn{2}{c}{2022} & \multicolumn{2}{c}{2023} & \multicolumn{2}{c}{2024} & \multicolumn{2}{c}{2025} & \multicolumn{2}{c}{2026} \\
\cmidrule(lr){3-4}\cmidrule(lr){5-6}\cmidrule(lr){7-8}\cmidrule(lr){9-10}\cmidrule(lr){11-12}\cmidrule(lr){13-14}
 & & OSSM & SSM & OSSM & SSM & OSSM & SSM & OSSM & SSM & OSSM & SSM & OSSM & SSM \\
\midrule
\endfirsthead
\multicolumn{14}{l}{\itshape Table \ref{tab:raw-cfg2-mse}, continued.}\\
\toprule
Dataset & $H$ & \multicolumn{2}{c}{2021} & \multicolumn{2}{c}{2022} & \multicolumn{2}{c}{2023} & \multicolumn{2}{c}{2024} & \multicolumn{2}{c}{2025} & \multicolumn{2}{c}{2026} \\
\cmidrule(lr){3-4}\cmidrule(lr){5-6}\cmidrule(lr){7-8}\cmidrule(lr){9-10}\cmidrule(lr){11-12}\cmidrule(lr){13-14}
 & & OSSM & SSM & OSSM & SSM & OSSM & SSM & OSSM & SSM & OSSM & SSM & OSSM & SSM \\
\midrule
\endhead
\endfoot
\endlastfoot
\multicolumn{14}{l}{\textit{Short-term forecasting}} \\*
\addlinespace[2pt]
NN5 & 24 & \textbf{1.156} & 1.313 & \textbf{1.152} & 1.322 & \textbf{1.196} & 1.272 & \textbf{1.096} & 1.277 & \textbf{1.126} & 1.365 & \textbf{1.198} & 1.281 \\*
 & 36 & \textbf{1.188} & 1.304 & \textbf{1.197} & 1.299 & \textbf{1.222} & 1.276 & \textbf{1.082} & 1.253 & \textbf{1.113} & 1.322 & \textbf{1.230} & 1.287 \\*
 & 48 & \textbf{1.205} & 1.296 & \textbf{1.226} & 1.290 & \textbf{1.226} & 1.264 & \textbf{1.067} & 1.239 & \textbf{1.103} & 1.308 & \textbf{1.235} & 1.269 \\*
 & 60 & \textbf{1.221} & 1.289 & \textbf{1.230} & 1.286 & \textbf{1.230} & 1.273 & \textbf{1.061} & 1.233 & \textbf{1.100} & 1.301 & \textbf{1.247} & 1.276 \\
\midrule
NYSE & 24 & 0.270 & \textbf{0.253} & 0.390 & \textbf{0.284} & 0.315 & \textbf{0.309} & 0.341 & \textbf{0.324} & 0.821 & \textbf{0.440} & 0.427 & \textbf{0.313} \\*
 & 36 & 0.678 & \textbf{0.415} & 1.346 & \textbf{1.014} & \textbf{0.468} & 0.605 & 0.607 & \textbf{0.532} & \textbf{0.532} & 0.712 & 0.704 & \textbf{0.406} \\*
 & 48 & 1.083 & \textbf{0.893} & 1.650 & \textbf{1.008} & 0.887 & \textbf{0.610} & \textbf{0.812} & 1.042 & \textbf{0.782} & 0.926 & 1.526 & \textbf{0.622} \\
\bottomrule
\end{longtable}
\endgroup


\begingroup
\setlength{\tabcolsep}{2pt}
\footnotesize
\begin{longtable}{l r cc cc cc cc cc cc }
\caption{Normalized MAE for configuration 2, lower is better. Rows are (dataset, forecast horizon $H$) pairs, columns are the six random seeds, each split into the observer head (OSSM) and the standard head (SSM). The better of the two is in bold. ``div.'' marks a run that diverged, i.e.\ reached a normalized MSE above 10 or produced no metric at all; divergence counts as a loss, so the head that converged is shown in bold.}\label{tab:raw-cfg2-mae}\\
\toprule
Dataset & $H$ & \multicolumn{2}{c}{2021} & \multicolumn{2}{c}{2022} & \multicolumn{2}{c}{2023} & \multicolumn{2}{c}{2024} & \multicolumn{2}{c}{2025} & \multicolumn{2}{c}{2026} \\
\cmidrule(lr){3-4}\cmidrule(lr){5-6}\cmidrule(lr){7-8}\cmidrule(lr){9-10}\cmidrule(lr){11-12}\cmidrule(lr){13-14}
 & & OSSM & SSM & OSSM & SSM & OSSM & SSM & OSSM & SSM & OSSM & SSM & OSSM & SSM \\
\midrule
\endfirsthead
\multicolumn{14}{l}{\itshape Table \ref{tab:raw-cfg2-mae}, continued.}\\
\toprule
Dataset & $H$ & \multicolumn{2}{c}{2021} & \multicolumn{2}{c}{2022} & \multicolumn{2}{c}{2023} & \multicolumn{2}{c}{2024} & \multicolumn{2}{c}{2025} & \multicolumn{2}{c}{2026} \\
\cmidrule(lr){3-4}\cmidrule(lr){5-6}\cmidrule(lr){7-8}\cmidrule(lr){9-10}\cmidrule(lr){11-12}\cmidrule(lr){13-14}
 & & OSSM & SSM & OSSM & SSM & OSSM & SSM & OSSM & SSM & OSSM & SSM & OSSM & SSM \\
\midrule
\endhead
\endfoot
\endlastfoot
\multicolumn{14}{l}{\textit{Short-term forecasting}} \\*
\addlinespace[2pt]
NN5 & 24 & \textbf{0.833} & 0.915 & \textbf{0.833} & 0.927 & \textbf{0.860} & 0.889 & \textbf{0.807} & 0.889 & \textbf{0.820} & 0.945 & \textbf{0.863} & 0.900 \\*
 & 36 & \textbf{0.862} & 0.924 & \textbf{0.869} & 0.925 & \textbf{0.886} & 0.909 & \textbf{0.809} & 0.887 & \textbf{0.824} & 0.934 & \textbf{0.891} & 0.917 \\*
 & 48 & \textbf{0.876} & 0.925 & \textbf{0.889} & 0.925 & \textbf{0.892} & 0.907 & \textbf{0.806} & 0.886 & \textbf{0.822} & 0.932 & \textbf{0.897} & 0.912 \\*
 & 60 & \textbf{0.887} & 0.924 & \textbf{0.892} & 0.926 & \textbf{0.896} & 0.916 & \textbf{0.805} & 0.884 & \textbf{0.823} & 0.932 & \textbf{0.906} & 0.919 \\
\midrule
NYSE & 24 & 0.360 & \textbf{0.359} & 0.456 & \textbf{0.383} & \textbf{0.394} & 0.400 & \textbf{0.411} & 0.415 & 0.667 & \textbf{0.462} & 0.465 & \textbf{0.400} \\*
 & 36 & 0.654 & \textbf{0.473} & 0.907 & \textbf{0.758} & \textbf{0.484} & 0.561 & 0.591 & \textbf{0.536} & \textbf{0.528} & 0.613 & 0.595 & \textbf{0.444} \\*
 & 48 & 0.849 & \textbf{0.695} & 1.019 & \textbf{0.742} & 0.691 & \textbf{0.572} & \textbf{0.688} & 0.791 & \textbf{0.638} & 0.706 & 0.964 & \textbf{0.588} \\
\bottomrule
\end{longtable}
\endgroup


\subsection{Configuration 3}

\begin{table}[H]
\centering
\caption{Hyper-parameters of configuration 3. Every other setting follows the
TFB defaults, and the two heads share all of them, including the random seeds.}
\small
\begin{tabular}{lc}
\toprule
Hyper-parameter & Value \\
\midrule
Encoder layers $e_{\mathrm{layers}}$ & 2 \\
State dimension $d_x$ & 128 \\
Model dimension $d_{\mathrm{model}}$ & 128 \\
Learning rate & $10^{-4}$ \\
Batch size & 32 \\
\bottomrule
\end{tabular}
\end{table}

\begingroup
\setlength{\tabcolsep}{2pt}
\footnotesize
\begin{longtable}{l r cc cc cc cc cc cc }
\caption{Normalized MSE for configuration 3, lower is better. Rows are (dataset, forecast horizon $H$) pairs, columns are the six random seeds, each split into the observer head (OSSM) and the standard head (SSM). The better of the two is in bold. ``div.'' marks a run that diverged, i.e.\ reached a normalized MSE above 10 or produced no metric at all; divergence counts as a loss, so the head that converged is shown in bold.}\label{tab:raw-cfg3-mse}\\
\toprule
Dataset & $H$ & \multicolumn{2}{c}{2021} & \multicolumn{2}{c}{2022} & \multicolumn{2}{c}{2023} & \multicolumn{2}{c}{2024} & \multicolumn{2}{c}{2025} & \multicolumn{2}{c}{2026} \\
\cmidrule(lr){3-4}\cmidrule(lr){5-6}\cmidrule(lr){7-8}\cmidrule(lr){9-10}\cmidrule(lr){11-12}\cmidrule(lr){13-14}
 & & OSSM & SSM & OSSM & SSM & OSSM & SSM & OSSM & SSM & OSSM & SSM & OSSM & SSM \\
\midrule
\endfirsthead
\multicolumn{14}{l}{\itshape Table \ref{tab:raw-cfg3-mse}, continued.}\\
\toprule
Dataset & $H$ & \multicolumn{2}{c}{2021} & \multicolumn{2}{c}{2022} & \multicolumn{2}{c}{2023} & \multicolumn{2}{c}{2024} & \multicolumn{2}{c}{2025} & \multicolumn{2}{c}{2026} \\
\cmidrule(lr){3-4}\cmidrule(lr){5-6}\cmidrule(lr){7-8}\cmidrule(lr){9-10}\cmidrule(lr){11-12}\cmidrule(lr){13-14}
 & & OSSM & SSM & OSSM & SSM & OSSM & SSM & OSSM & SSM & OSSM & SSM & OSSM & SSM \\
\midrule
\endhead
\endfoot
\endlastfoot
\multicolumn{14}{l}{\textit{Short-term forecasting}} \\*
\addlinespace[2pt]
NN5 & 24 & \textbf{1.020} & 1.246 & \textbf{0.993} & 1.150 & \textbf{0.996} & 1.324 & \textbf{1.015} & 1.314 & \textbf{1.022} & 1.225 & \textbf{1.058} & 1.272 \\*
 & 36 & \textbf{1.014} & 1.233 & \textbf{0.980} & 1.154 & \textbf{0.969} & 1.310 & \textbf{0.997} & 1.302 & \textbf{1.004} & 1.203 & \textbf{1.047} & 1.237 \\*
 & 48 & \textbf{0.986} & 1.227 & \textbf{0.973} & 1.132 & \textbf{0.964} & 1.297 & \textbf{0.984} & 1.292 & \textbf{0.997} & 1.183 & \textbf{1.035} & 1.226 \\*
 & 60 & \textbf{0.977} & 1.222 & \textbf{0.967} & 1.138 & \textbf{0.953} & 1.294 & \textbf{0.983} & 1.288 & \textbf{0.984} & 1.178 & \textbf{1.022} & 1.235 \\
\midrule
NYSE & 24 & \textbf{0.265} & 0.305 & \textbf{0.236} & 0.261 & 0.246 & \textbf{0.244} & \textbf{0.261} & 0.300 & \textbf{0.250} & 0.262 & \textbf{0.258} & 0.258 \\*
 & 36 & \textbf{0.352} & 0.442 & \textbf{0.416} & 0.445 & \textbf{0.536} & 0.545 & \textbf{0.443} & 0.514 & \textbf{0.474} & 0.477 & \textbf{0.404} & 0.647 \\*
 & 48 & \textbf{0.566} & 0.602 & \textbf{1.036} & 1.220 & 1.004 & \textbf{0.779} & \textbf{0.760} & 1.366 & 0.732 & \textbf{0.633} & 0.992 & \textbf{0.653} \\
\bottomrule
\end{longtable}
\endgroup


\begingroup
\setlength{\tabcolsep}{2pt}
\footnotesize
\begin{longtable}{l r cc cc cc cc cc cc }
\caption{Normalized MAE for configuration 3, lower is better. Rows are (dataset, forecast horizon $H$) pairs, columns are the six random seeds, each split into the observer head (OSSM) and the standard head (SSM). The better of the two is in bold. ``div.'' marks a run that diverged, i.e.\ reached a normalized MSE above 10 or produced no metric at all; divergence counts as a loss, so the head that converged is shown in bold.}\label{tab:raw-cfg3-mae}\\
\toprule
Dataset & $H$ & \multicolumn{2}{c}{2021} & \multicolumn{2}{c}{2022} & \multicolumn{2}{c}{2023} & \multicolumn{2}{c}{2024} & \multicolumn{2}{c}{2025} & \multicolumn{2}{c}{2026} \\
\cmidrule(lr){3-4}\cmidrule(lr){5-6}\cmidrule(lr){7-8}\cmidrule(lr){9-10}\cmidrule(lr){11-12}\cmidrule(lr){13-14}
 & & OSSM & SSM & OSSM & SSM & OSSM & SSM & OSSM & SSM & OSSM & SSM & OSSM & SSM \\
\midrule
\endfirsthead
\multicolumn{14}{l}{\itshape Table \ref{tab:raw-cfg3-mae}, continued.}\\
\toprule
Dataset & $H$ & \multicolumn{2}{c}{2021} & \multicolumn{2}{c}{2022} & \multicolumn{2}{c}{2023} & \multicolumn{2}{c}{2024} & \multicolumn{2}{c}{2025} & \multicolumn{2}{c}{2026} \\
\cmidrule(lr){3-4}\cmidrule(lr){5-6}\cmidrule(lr){7-8}\cmidrule(lr){9-10}\cmidrule(lr){11-12}\cmidrule(lr){13-14}
 & & OSSM & SSM & OSSM & SSM & OSSM & SSM & OSSM & SSM & OSSM & SSM & OSSM & SSM \\
\midrule
\endhead
\endfoot
\endlastfoot
\multicolumn{14}{l}{\textit{Short-term forecasting}} \\*
\addlinespace[2pt]
NN5 & 24 & \textbf{0.759} & 0.872 & \textbf{0.749} & 0.827 & \textbf{0.747} & 0.924 & \textbf{0.759} & 0.918 & \textbf{0.760} & 0.856 & \textbf{0.786} & 0.881 \\*
 & 36 & \textbf{0.767} & 0.877 & \textbf{0.753} & 0.836 & \textbf{0.743} & 0.928 & \textbf{0.760} & 0.924 & \textbf{0.763} & 0.855 & \textbf{0.791} & 0.875 \\*
 & 48 & \textbf{0.758} & 0.879 & \textbf{0.754} & 0.830 & \textbf{0.746} & 0.926 & \textbf{0.758} & 0.924 & \textbf{0.765} & 0.850 & \textbf{0.789} & 0.874 \\*
 & 60 & \textbf{0.755} & 0.879 & \textbf{0.753} & 0.834 & \textbf{0.743} & 0.928 & \textbf{0.760} & 0.925 & \textbf{0.762} & 0.850 & \textbf{0.784} & 0.882 \\
\midrule
NYSE & 24 & \textbf{0.355} & 0.402 & \textbf{0.337} & 0.364 & 0.340 & \textbf{0.337} & \textbf{0.373} & 0.388 & \textbf{0.356} & 0.358 & \textbf{0.348} & 0.359 \\*
 & 36 & \textbf{0.394} & 0.497 & \textbf{0.457} & 0.480 & 0.526 & \textbf{0.522} & \textbf{0.477} & 0.526 & \textbf{0.507} & 0.511 & \textbf{0.436} & 0.599 \\*
 & 48 & \textbf{0.529} & 0.594 & \textbf{0.743} & 0.834 & 0.782 & \textbf{0.633} & \textbf{0.637} & 0.911 & 0.624 & \textbf{0.581} & 0.732 & \textbf{0.562} \\
\bottomrule
\end{longtable}
\endgroup


\subsection{Configuration 4}

\begin{table}[H]
\centering
\caption{Hyper-parameters of configuration 4. Every other setting follows the
TFB defaults, and the two heads share all of them, including the random seeds.}
\small
\begin{tabular}{lc}
\toprule
Hyper-parameter & Value \\
\midrule
Encoder layers $e_{\mathrm{layers}}$ & 2 \\
State dimension $d_x$ & 128 \\
Model dimension $d_{\mathrm{model}}$ & 128 \\
Learning rate & $10^{-3}$ \\
Batch size & 32 \\
\bottomrule
\end{tabular}
\end{table}

\begingroup
\setlength{\tabcolsep}{2pt}
\footnotesize
\begin{longtable}{l r cc cc cc cc cc cc }
\caption{Normalized MSE for configuration 4, lower is better. Rows are (dataset, forecast horizon $H$) pairs, columns are the six random seeds, each split into the observer head (OSSM) and the standard head (SSM). The better of the two is in bold. ``div.'' marks a run that diverged, i.e.\ reached a normalized MSE above 10 or produced no metric at all; divergence counts as a loss, so the head that converged is shown in bold.}\label{tab:raw-cfg4-mse}\\
\toprule
Dataset & $H$ & \multicolumn{2}{c}{2021} & \multicolumn{2}{c}{2022} & \multicolumn{2}{c}{2023} & \multicolumn{2}{c}{2024} & \multicolumn{2}{c}{2025} & \multicolumn{2}{c}{2026} \\
\cmidrule(lr){3-4}\cmidrule(lr){5-6}\cmidrule(lr){7-8}\cmidrule(lr){9-10}\cmidrule(lr){11-12}\cmidrule(lr){13-14}
 & & OSSM & SSM & OSSM & SSM & OSSM & SSM & OSSM & SSM & OSSM & SSM & OSSM & SSM \\
\midrule
\endfirsthead
\multicolumn{14}{l}{\itshape Table \ref{tab:raw-cfg4-mse}, continued.}\\
\toprule
Dataset & $H$ & \multicolumn{2}{c}{2021} & \multicolumn{2}{c}{2022} & \multicolumn{2}{c}{2023} & \multicolumn{2}{c}{2024} & \multicolumn{2}{c}{2025} & \multicolumn{2}{c}{2026} \\
\cmidrule(lr){3-4}\cmidrule(lr){5-6}\cmidrule(lr){7-8}\cmidrule(lr){9-10}\cmidrule(lr){11-12}\cmidrule(lr){13-14}
 & & OSSM & SSM & OSSM & SSM & OSSM & SSM & OSSM & SSM & OSSM & SSM & OSSM & SSM \\
\midrule
\endhead
\endfoot
\endlastfoot
\multicolumn{14}{l}{\textit{Short-term forecasting}} \\*
\addlinespace[2pt]
NN5 & 24 & \textbf{0.800} & 1.071 & \textbf{0.759} & 0.854 & 0.836 & \textbf{0.823} & \textbf{0.810} & 1.316 & \textbf{0.812} & 1.198 & \textbf{0.807} & 0.862 \\*
 & 36 & \textbf{0.843} & 0.947 & \textbf{0.771} & 1.165 & \textbf{1.107} & 1.300 & \textbf{0.819} & 0.823 & \textbf{0.850} & 1.300 & \textbf{0.757} & 1.193 \\*
 & 48 & \textbf{1.090} & 1.299 & \textbf{0.795} & 1.306 & \textbf{0.734} & 1.241 & \textbf{1.190} & 1.282 & \textbf{0.757} & 0.842 & \textbf{1.107} & 1.299 \\*
 & 60 & 1.265 & \textbf{1.179} & \textbf{0.720} & 1.290 & \textbf{1.264} & 1.290 & \textbf{0.764} & 1.291 & 1.274 & \textbf{1.267} & \textbf{0.905} & 1.226 \\
\midrule
NYSE & 24 & \textbf{0.282} & 0.591 & 0.503 & \textbf{0.227} & \textbf{0.228} & 0.283 & 0.275 & \textbf{0.275} & \textbf{0.209} & 0.243 & 0.354 & \textbf{0.294} \\*
 & 36 & \textbf{0.397} & 1.471 & 1.365 & \textbf{0.906} & \textbf{1.444} & 1.622 & \textbf{0.367} & 1.338 & \textbf{0.436} & 1.459 & \textbf{0.438} & 1.479 \\*
 & 48 & \textbf{0.543} & 1.797 & \textbf{0.560} & 1.725 & \textbf{1.738} & 1.761 & \textbf{1.670} & 1.726 & \textbf{1.435} & 1.735 & \textbf{1.519} & 1.772 \\
\bottomrule
\end{longtable}
\endgroup


\begingroup
\setlength{\tabcolsep}{2pt}
\footnotesize
\begin{longtable}{l r cc cc cc cc cc cc }
\caption{Normalized MAE for configuration 4, lower is better. Rows are (dataset, forecast horizon $H$) pairs, columns are the six random seeds, each split into the observer head (OSSM) and the standard head (SSM). The better of the two is in bold. ``div.'' marks a run that diverged, i.e.\ reached a normalized MSE above 10 or produced no metric at all; divergence counts as a loss, so the head that converged is shown in bold.}\label{tab:raw-cfg4-mae}\\
\toprule
Dataset & $H$ & \multicolumn{2}{c}{2021} & \multicolumn{2}{c}{2022} & \multicolumn{2}{c}{2023} & \multicolumn{2}{c}{2024} & \multicolumn{2}{c}{2025} & \multicolumn{2}{c}{2026} \\
\cmidrule(lr){3-4}\cmidrule(lr){5-6}\cmidrule(lr){7-8}\cmidrule(lr){9-10}\cmidrule(lr){11-12}\cmidrule(lr){13-14}
 & & OSSM & SSM & OSSM & SSM & OSSM & SSM & OSSM & SSM & OSSM & SSM & OSSM & SSM \\
\midrule
\endfirsthead
\multicolumn{14}{l}{\itshape Table \ref{tab:raw-cfg4-mae}, continued.}\\
\toprule
Dataset & $H$ & \multicolumn{2}{c}{2021} & \multicolumn{2}{c}{2022} & \multicolumn{2}{c}{2023} & \multicolumn{2}{c}{2024} & \multicolumn{2}{c}{2025} & \multicolumn{2}{c}{2026} \\
\cmidrule(lr){3-4}\cmidrule(lr){5-6}\cmidrule(lr){7-8}\cmidrule(lr){9-10}\cmidrule(lr){11-12}\cmidrule(lr){13-14}
 & & OSSM & SSM & OSSM & SSM & OSSM & SSM & OSSM & SSM & OSSM & SSM & OSSM & SSM \\
\midrule
\endhead
\endfoot
\endlastfoot
\multicolumn{14}{l}{\textit{Short-term forecasting}} \\*
\addlinespace[2pt]
NN5 & 24 & \textbf{0.628} & 0.783 & \textbf{0.606} & 0.663 & \textbf{0.635} & 0.644 & \textbf{0.629} & 0.929 & \textbf{0.624} & 0.870 & \textbf{0.627} & 0.658 \\*
 & 36 & \textbf{0.682} & 0.722 & \textbf{0.623} & 0.859 & \textbf{0.834} & 0.926 & \textbf{0.637} & 0.649 & \textbf{0.678} & 0.927 & \textbf{0.618} & 0.879 \\*
 & 48 & \textbf{0.824} & 0.917 & \textbf{0.657} & 0.932 & \textbf{0.605} & 0.902 & \textbf{0.873} & 0.925 & \textbf{0.616} & 0.663 & \textbf{0.829} & 0.930 \\*
 & 60 & 0.920 & \textbf{0.868} & \textbf{0.606} & 0.930 & \textbf{0.918} & 0.928 & \textbf{0.623} & 0.929 & 0.923 & \textbf{0.918} & \textbf{0.711} & 0.899 \\
\midrule
NYSE & 24 & \textbf{0.379} & 0.554 & 0.507 & \textbf{0.325} & \textbf{0.331} & 0.376 & \textbf{0.361} & 0.373 & \textbf{0.307} & 0.342 & 0.428 & \textbf{0.388} \\*
 & 36 & \textbf{0.445} & 0.968 & 0.923 & \textbf{0.701} & \textbf{0.955} & 0.993 & \textbf{0.419} & 0.913 & \textbf{0.469} & 0.962 & \textbf{0.476} & 0.964 \\*
 & 48 & \textbf{0.517} & 1.083 & \textbf{0.528} & 1.057 & \textbf{1.065} & 1.071 & \textbf{1.043} & 1.060 & \textbf{0.924} & 1.062 & \textbf{0.963} & 1.074 \\
\bottomrule
\end{longtable}
\endgroup


\subsection{Configuration 5}

\begin{table}[H]
\centering
\caption{Hyper-parameters of configuration 5. Every other setting follows the
TFB defaults, and the two heads share all of them, including the random seeds.}
\small
\begin{tabular}{lc}
\toprule
Hyper-parameter & Value \\
\midrule
Encoder layers $e_{\mathrm{layers}}$ & 2 \\
State dimension $d_x$ & 128 \\
Model dimension $d_{\mathrm{model}}$ & 128 \\
Learning rate & $10^{-3}$ \\
Batch size & 64 \\
\bottomrule
\end{tabular}
\end{table}

\begingroup
\setlength{\tabcolsep}{2pt}
\footnotesize
\begin{longtable}{l r cc cc cc cc cc cc }
\caption{normalized MSE for configuration 5, lower is better. Rows are (dataset, forecast horizon $H$) pairs, columns are the six random seeds, each split into the observer head (OSSM) and the standard head (SSM). The better of the two is in bold. ``div.'' marks a run that diverged, i.e.\ reached a normalized MSE above 10 or produced no metric at all; divergence counts as a loss, so the head that converged is shown in bold.}\label{tab:raw-cfg5-mse}\\
\toprule
Dataset & $H$ & \multicolumn{2}{c}{2021} & \multicolumn{2}{c}{2022} & \multicolumn{2}{c}{2023} & \multicolumn{2}{c}{2024} & \multicolumn{2}{c}{2025} & \multicolumn{2}{c}{2026} \\
\cmidrule(lr){3-4}\cmidrule(lr){5-6}\cmidrule(lr){7-8}\cmidrule(lr){9-10}\cmidrule(lr){11-12}\cmidrule(lr){13-14}
 & & OSSM & SSM & OSSM & SSM & OSSM & SSM & OSSM & SSM & OSSM & SSM & OSSM & SSM \\
\midrule
\endfirsthead
\multicolumn{14}{l}{\itshape Table \ref{tab:raw-cfg5-mse}, continued.}\\
\toprule
Dataset & $H$ & \multicolumn{2}{c}{2021} & \multicolumn{2}{c}{2022} & \multicolumn{2}{c}{2023} & \multicolumn{2}{c}{2024} & \multicolumn{2}{c}{2025} & \multicolumn{2}{c}{2026} \\
\cmidrule(lr){3-4}\cmidrule(lr){5-6}\cmidrule(lr){7-8}\cmidrule(lr){9-10}\cmidrule(lr){11-12}\cmidrule(lr){13-14}
 & & OSSM & SSM & OSSM & SSM & OSSM & SSM & OSSM & SSM & OSSM & SSM & OSSM & SSM \\
\midrule
\endhead
\endfoot
\endlastfoot
\multicolumn{14}{l}{\textit{Short-term forecasting}} \\*
\addlinespace[2pt]
NN5 & 24 & \textbf{1.104} & 1.123 & \textbf{1.215} & 1.345 & \textbf{0.836} & 0.925 & 1.127 & \textbf{1.001} & \textbf{0.949} & 1.299 & \textbf{0.875} & 1.263 \\*
 & 36 & \textbf{1.156} & 1.298 & \textbf{1.157} & 1.284 & \textbf{0.863} & 1.295 & \textbf{0.896} & 1.295 & \textbf{1.161} & 1.327 & \textbf{0.780} & 1.254 \\*
 & 48 & \textbf{0.964} & 1.279 & \textbf{0.751} & 1.301 & \textbf{0.803} & 1.281 & \textbf{1.068} & 1.288 & \textbf{1.247} & 1.315 & \textbf{0.783} & 1.289 \\*
 & 60 & \textbf{1.262} & 1.289 & \textbf{1.211} & 1.296 & \textbf{1.115} & 1.286 & \textbf{1.133} & 1.287 & \textbf{1.281} & 1.291 & \textbf{0.766} & 1.280 \\
\midrule
NYSE & 24 & \textbf{0.369} & 1.116 & \textbf{0.239} & 1.210 & \textbf{0.659} & 1.181 & \textbf{0.221} & 0.317 & \textbf{0.308} & 1.064 & \textbf{0.297} & 1.178 \\*
 & 36 & \textbf{1.474} & 1.535 & \textbf{1.249} & 1.486 & \textbf{1.470} & 1.481 & \textbf{0.466} & 1.362 & \textbf{1.363} & 1.473 & \textbf{1.226} & 1.474 \\*
 & 48 & \textbf{1.626} & 1.804 & \textbf{1.698} & 1.709 & \textbf{0.767} & 1.758 & \textbf{0.797} & 1.741 & \textbf{0.591} & 1.744 & \textbf{1.472} & 1.752 \\
\bottomrule
\end{longtable}
\endgroup


\begingroup
\setlength{\tabcolsep}{2pt}
\footnotesize
\begin{longtable}{l r cc cc cc cc cc cc }
\caption{normalized MAE for configuration 5, lower is better. Rows are (dataset, forecast horizon $H$) pairs, columns are the six random seeds, each split into the observer head (OSSM) and the standard head (SSM). The better of the two is in bold. ``div.'' marks a run that diverged, i.e.\ reached a normalized MSE above 10 or produced no metric at all; divergence counts as a loss, so the head that converged is shown in bold.}\label{tab:raw-cfg5-mae}\\
\toprule
Dataset & $H$ & \multicolumn{2}{c}{2021} & \multicolumn{2}{c}{2022} & \multicolumn{2}{c}{2023} & \multicolumn{2}{c}{2024} & \multicolumn{2}{c}{2025} & \multicolumn{2}{c}{2026} \\
\cmidrule(lr){3-4}\cmidrule(lr){5-6}\cmidrule(lr){7-8}\cmidrule(lr){9-10}\cmidrule(lr){11-12}\cmidrule(lr){13-14}
 & & OSSM & SSM & OSSM & SSM & OSSM & SSM & OSSM & SSM & OSSM & SSM & OSSM & SSM \\
\midrule
\endfirsthead
\multicolumn{14}{l}{\itshape Table \ref{tab:raw-cfg5-mae}, continued.}\\
\toprule
Dataset & $H$ & \multicolumn{2}{c}{2021} & \multicolumn{2}{c}{2022} & \multicolumn{2}{c}{2023} & \multicolumn{2}{c}{2024} & \multicolumn{2}{c}{2025} & \multicolumn{2}{c}{2026} \\
\cmidrule(lr){3-4}\cmidrule(lr){5-6}\cmidrule(lr){7-8}\cmidrule(lr){9-10}\cmidrule(lr){11-12}\cmidrule(lr){13-14}
 & & OSSM & SSM & OSSM & SSM & OSSM & SSM & OSSM & SSM & OSSM & SSM & OSSM & SSM \\
\midrule
\endhead
\endfoot
\endlastfoot
\multicolumn{14}{l}{\textit{Short-term forecasting}} \\*
\addlinespace[2pt]
NN5 & 24 & \textbf{0.804} & 0.814 & \textbf{0.883} & 0.940 & \textbf{0.642} & 0.701 & 0.838 & \textbf{0.747} & \textbf{0.703} & 0.918 & \textbf{0.670} & 0.904 \\*
 & 36 & \textbf{0.851} & 0.924 & \textbf{0.860} & 0.921 & \textbf{0.660} & 0.924 & \textbf{0.696} & 0.925 & \textbf{0.852} & 0.937 & \textbf{0.631} & 0.906 \\*
 & 48 & \textbf{0.740} & 0.919 & \textbf{0.628} & 0.930 & \textbf{0.641} & 0.922 & \textbf{0.813} & 0.924 & \textbf{0.904} & 0.935 & \textbf{0.638} & 0.923 \\*
 & 60 & \textbf{0.911} & 0.927 & \textbf{0.890} & 0.931 & \textbf{0.835} & 0.927 & \textbf{0.857} & 0.926 & \textbf{0.925} & 0.929 & \textbf{0.634} & 0.922 \\
\midrule
NYSE & 24 & \textbf{0.436} & 0.821 & \textbf{0.334} & 0.862 & \textbf{0.602} & 0.849 & \textbf{0.324} & 0.399 & \textbf{0.393} & 0.801 & \textbf{0.399} & 0.852 \\*
 & 36 & \textbf{0.968} & 0.989 & \textbf{0.879} & 0.969 & \textbf{0.967} & 0.970 & \textbf{0.487} & 0.923 & \textbf{0.925} & 0.968 & \textbf{0.868} & 0.969 \\*
 & 48 & \textbf{1.016} & 1.086 & \textbf{1.049} & 1.050 & \textbf{0.631} & 1.069 & \textbf{0.637} & 1.065 & \textbf{0.558} & 1.064 & \textbf{0.934} & 1.068 \\
\bottomrule
\end{longtable}
\endgroup


\subsection{Configuration 6}

\begin{table}[H]
\centering
\caption{Hyper-parameters of configuration 6. Every other setting follows the
TFB defaults, and the two heads share all of them, including the random seeds.}
\small
\begin{tabular}{lc}
\toprule
Hyper-parameter & Value \\
\midrule
Encoder layers $e_{\mathrm{layers}}$ & 2 \\
State dimension $d_x$ & 256 \\
Model dimension $d_{\mathrm{model}}$ & 128 \\
Learning rate & $10^{-4}$ \\
Batch size & 32 \\
\bottomrule
\end{tabular}
\end{table}

\begingroup
\setlength{\tabcolsep}{2pt}
\footnotesize
\begin{longtable}{l r cc cc cc cc cc cc }
\caption{normalized MSE for configuration 6, lower is better. Rows are (dataset, forecast horizon $H$) pairs, columns are the six random seeds, each split into the observer head (OSSM) and the standard head (SSM). The better of the two is in bold. ``div.'' marks a run that diverged, i.e.\ reached a normalized MSE above 10 or produced no metric at all; divergence counts as a loss, so the head that converged is shown in bold.}\label{tab:raw-cfg6-mse}\\
\toprule
Dataset & $H$ & \multicolumn{2}{c}{2021} & \multicolumn{2}{c}{2022} & \multicolumn{2}{c}{2023} & \multicolumn{2}{c}{2024} & \multicolumn{2}{c}{2025} & \multicolumn{2}{c}{2026} \\
\cmidrule(lr){3-4}\cmidrule(lr){5-6}\cmidrule(lr){7-8}\cmidrule(lr){9-10}\cmidrule(lr){11-12}\cmidrule(lr){13-14}
 & & OSSM & SSM & OSSM & SSM & OSSM & SSM & OSSM & SSM & OSSM & SSM & OSSM & SSM \\
\midrule
\endfirsthead
\multicolumn{14}{l}{\itshape Table \ref{tab:raw-cfg6-mse}, continued.}\\
\toprule
Dataset & $H$ & \multicolumn{2}{c}{2021} & \multicolumn{2}{c}{2022} & \multicolumn{2}{c}{2023} & \multicolumn{2}{c}{2024} & \multicolumn{2}{c}{2025} & \multicolumn{2}{c}{2026} \\
\cmidrule(lr){3-4}\cmidrule(lr){5-6}\cmidrule(lr){7-8}\cmidrule(lr){9-10}\cmidrule(lr){11-12}\cmidrule(lr){13-14}
 & & OSSM & SSM & OSSM & SSM & OSSM & SSM & OSSM & SSM & OSSM & SSM & OSSM & SSM \\
\midrule
\endhead
\endfoot
\endlastfoot
\multicolumn{14}{l}{\textit{Short-term forecasting}} \\*
\addlinespace[2pt]
NN5 & 24 & \textbf{0.939} & 1.094 & \textbf{0.876} & 1.118 & \textbf{0.901} & 1.079 & \textbf{0.927} & 1.133 & \textbf{0.923} & 1.128 & \textbf{0.894} & 1.123 \\*
 & 36 & \textbf{0.893} & 1.105 & \textbf{0.855} & 1.097 & \textbf{0.880} & 1.082 & \textbf{0.887} & 1.122 & \textbf{0.902} & 1.112 & \textbf{0.893} & 1.118 \\*
 & 48 & \textbf{0.886} & 1.097 & \textbf{0.831} & 1.077 & \textbf{0.848} & 1.056 & \textbf{0.870} & 1.115 & \textbf{0.878} & 1.110 & \textbf{0.866} & 1.119 \\*
 & 60 & \textbf{0.877} & 1.115 & \textbf{0.812} & 1.085 & \textbf{0.897} & 1.112 & \textbf{0.859} & 1.146 & \textbf{0.873} & 1.144 & \textbf{0.861} & 1.128 \\
\midrule
NYSE & 24 & \textbf{0.251} & 0.268 & 0.325 & \textbf{0.321} & 0.523 & \textbf{0.270} & \textbf{0.251} & 0.374 & 0.539 & \textbf{0.270} & \textbf{0.251} & 0.274 \\*
 & 36 & \textbf{0.378} & 0.743 & \textbf{0.408} & 0.710 & \textbf{0.412} & 0.421 & 0.583 & \textbf{0.441} & \textbf{0.385} & 0.744 & 0.510 & \textbf{0.465} \\*
 & 48 & \textbf{1.390} & 1.511 & \textbf{0.886} & 1.462 & \textbf{0.925} & 1.584 & \textbf{0.759} & 1.592 & \textbf{0.949} & 1.273 & \textbf{0.589} & 0.652 \\
\bottomrule
\end{longtable}
\endgroup


\begingroup
\setlength{\tabcolsep}{2pt}
\footnotesize
\begin{longtable}{l r cc cc cc cc cc cc }
\caption{normalized MAE for configuration 6, lower is better. Rows are (dataset, forecast horizon $H$) pairs, columns are the six random seeds, each split into the observer head (OSSM) and the standard head (SSM). The better of the two is in bold. ``div.'' marks a run that diverged, i.e.\ reached a normalized MSE above 10 or produced no metric at all; divergence counts as a loss, so the head that converged is shown in bold.}\label{tab:raw-cfg6-mae}\\
\toprule
Dataset & $H$ & \multicolumn{2}{c}{2021} & \multicolumn{2}{c}{2022} & \multicolumn{2}{c}{2023} & \multicolumn{2}{c}{2024} & \multicolumn{2}{c}{2025} & \multicolumn{2}{c}{2026} \\
\cmidrule(lr){3-4}\cmidrule(lr){5-6}\cmidrule(lr){7-8}\cmidrule(lr){9-10}\cmidrule(lr){11-12}\cmidrule(lr){13-14}
 & & OSSM & SSM & OSSM & SSM & OSSM & SSM & OSSM & SSM & OSSM & SSM & OSSM & SSM \\
\midrule
\endfirsthead
\multicolumn{14}{l}{\itshape Table \ref{tab:raw-cfg6-mae}, continued.}\\
\toprule
Dataset & $H$ & \multicolumn{2}{c}{2021} & \multicolumn{2}{c}{2022} & \multicolumn{2}{c}{2023} & \multicolumn{2}{c}{2024} & \multicolumn{2}{c}{2025} & \multicolumn{2}{c}{2026} \\
\cmidrule(lr){3-4}\cmidrule(lr){5-6}\cmidrule(lr){7-8}\cmidrule(lr){9-10}\cmidrule(lr){11-12}\cmidrule(lr){13-14}
 & & OSSM & SSM & OSSM & SSM & OSSM & SSM & OSSM & SSM & OSSM & SSM & OSSM & SSM \\
\midrule
\endhead
\endfoot
\endlastfoot
\multicolumn{14}{l}{\textit{Short-term forecasting}} \\*
\addlinespace[2pt]
NN5 & 24 & \textbf{0.719} & 0.796 & \textbf{0.679} & 0.804 & \textbf{0.693} & 0.780 & \textbf{0.715} & 0.811 & \textbf{0.703} & 0.810 & \textbf{0.692} & 0.813 \\*
 & 36 & \textbf{0.707} & 0.809 & \textbf{0.683} & 0.802 & \textbf{0.700} & 0.794 & \textbf{0.707} & 0.813 & \textbf{0.710} & 0.810 & \textbf{0.705} & 0.820 \\*
 & 48 & \textbf{0.710} & 0.809 & \textbf{0.676} & 0.796 & \textbf{0.688} & 0.786 & \textbf{0.703} & 0.813 & \textbf{0.703} & 0.812 & \textbf{0.698} & 0.825 \\*
 & 60 & \textbf{0.707} & 0.820 & \textbf{0.668} & 0.801 & \textbf{0.718} & 0.818 & \textbf{0.698} & 0.831 & \textbf{0.704} & 0.831 & \textbf{0.697} & 0.831 \\
\midrule
NYSE & 24 & \textbf{0.337} & 0.363 & 0.412 & \textbf{0.406} & 0.505 & \textbf{0.365} & \textbf{0.348} & 0.449 & 0.526 & \textbf{0.367} & \textbf{0.337} & 0.362 \\*
 & 36 & \textbf{0.427} & 0.619 & \textbf{0.456} & 0.623 & \textbf{0.454} & 0.460 & 0.576 & \textbf{0.483} & \textbf{0.432} & 0.643 & 0.513 & \textbf{0.490} \\*
 & 48 & \textbf{0.909} & 0.953 & \textbf{0.700} & 0.935 & \textbf{0.707} & 0.993 & \textbf{0.652} & 1.001 & \textbf{0.710} & 0.864 & \textbf{0.556} & 0.588 \\
\bottomrule
\end{longtable}
\endgroup


\subsection{Configuration 7}

\begin{table}[H]
\centering
\caption{Hyper-parameters of configuration 7. Every other setting follows the
TFB defaults, and the two heads share all of them, including the random seeds.}
\small
\begin{tabular}{lc}
\toprule
Hyper-parameter & Value \\
\midrule
Encoder layers $e_{\mathrm{layers}}$ & 2 \\
State dimension $d_x$ & 256 \\
Model dimension $d_{\mathrm{model}}$ & 256 \\
Learning rate & $10^{-4}$ \\
Batch size & 32 \\
\bottomrule
\end{tabular}
\end{table}

\begingroup
\setlength{\tabcolsep}{2pt}
\footnotesize
\begin{longtable}{l r cc cc cc cc cc cc }
\caption{normalized MSE for configuration 7, lower is better. Rows are (dataset, forecast horizon $H$) pairs, columns are the six random seeds, each split into the observer head (OSSM) and the standard head (SSM). The better of the two is in bold. ``div.'' marks a run that diverged, i.e.\ reached a normalized MSE above 10 or produced no metric at all; divergence counts as a loss, so the head that converged is shown in bold.}\label{tab:raw-cfg7-mse}\\
\toprule
Dataset & $H$ & \multicolumn{2}{c}{2021} & \multicolumn{2}{c}{2022} & \multicolumn{2}{c}{2023} & \multicolumn{2}{c}{2024} & \multicolumn{2}{c}{2025} & \multicolumn{2}{c}{2026} \\
\cmidrule(lr){3-4}\cmidrule(lr){5-6}\cmidrule(lr){7-8}\cmidrule(lr){9-10}\cmidrule(lr){11-12}\cmidrule(lr){13-14}
 & & OSSM & SSM & OSSM & SSM & OSSM & SSM & OSSM & SSM & OSSM & SSM & OSSM & SSM \\
\midrule
\endfirsthead
\multicolumn{14}{l}{\itshape Table \ref{tab:raw-cfg7-mse}, continued.}\\
\toprule
Dataset & $H$ & \multicolumn{2}{c}{2021} & \multicolumn{2}{c}{2022} & \multicolumn{2}{c}{2023} & \multicolumn{2}{c}{2024} & \multicolumn{2}{c}{2025} & \multicolumn{2}{c}{2026} \\
\cmidrule(lr){3-4}\cmidrule(lr){5-6}\cmidrule(lr){7-8}\cmidrule(lr){9-10}\cmidrule(lr){11-12}\cmidrule(lr){13-14}
 & & OSSM & SSM & OSSM & SSM & OSSM & SSM & OSSM & SSM & OSSM & SSM & OSSM & SSM \\
\midrule
\endhead
\endfoot
\endlastfoot
\multicolumn{14}{l}{\textit{Short-term forecasting}} \\*
\addlinespace[2pt]
NN5 & 24 & \textbf{0.813} & 0.971 & \textbf{0.832} & 1.027 & \textbf{0.877} & 1.081 & \textbf{0.802} & 1.022 & \textbf{0.866} & 0.979 & \textbf{0.834} & 1.021 \\*
 & 36 & \textbf{0.770} & 0.933 & \textbf{0.810} & 1.031 & \textbf{0.872} & 1.056 & \textbf{0.770} & 0.979 & \textbf{0.830} & 0.995 & \textbf{0.828} & 0.992 \\*
 & 48 & \textbf{0.752} & 1.167 & \textbf{0.801} & 1.055 & \textbf{0.836} & 1.035 & \textbf{0.749} & 0.985 & \textbf{0.809} & 1.029 & \textbf{0.802} & 1.236 \\*
 & 60 & \textbf{0.723} & 0.908 & \textbf{0.814} & 1.011 & \textbf{0.830} & 1.083 & \textbf{0.793} & 1.080 & \textbf{0.798} & 1.020 & \textbf{0.803} & 1.237 \\
\midrule
NYSE & 24 & 0.263 & \textbf{0.254} & \textbf{0.228} & 0.244 & \textbf{0.240} & 0.282 & 0.238 & \textbf{0.228} & 0.296 & \textbf{0.257} & \textbf{0.235} & 0.866 \\*
 & 36 & 0.485 & \textbf{0.435} & 0.513 & \textbf{0.504} & \textbf{0.398} & 0.542 & \textbf{0.491} & 0.870 & \textbf{0.568} & 1.165 & \textbf{0.771} & 0.953 \\*
 & 48 & 0.862 & \textbf{0.742} & \textbf{1.019} & 1.597 & \textbf{0.788} & 1.493 & \textbf{0.893} & 0.955 & \textbf{0.763} & 1.551 & \textbf{0.469} & 1.624 \\
\bottomrule
\end{longtable}
\endgroup


\begingroup
\setlength{\tabcolsep}{2pt}
\footnotesize
\begin{longtable}{l r cc cc cc cc cc cc }
\caption{normalized MAE for configuration 7, lower is better. Rows are (dataset, forecast horizon $H$) pairs, columns are the six random seeds, each split into the observer head (OSSM) and the standard head (SSM). The better of the two is in bold. ``div.'' marks a run that diverged, i.e.\ reached a normalized MSE above 10 or produced no metric at all; divergence counts as a loss, so the head that converged is shown in bold.}\label{tab:raw-cfg7-mae}\\
\toprule
Dataset & $H$ & \multicolumn{2}{c}{2021} & \multicolumn{2}{c}{2022} & \multicolumn{2}{c}{2023} & \multicolumn{2}{c}{2024} & \multicolumn{2}{c}{2025} & \multicolumn{2}{c}{2026} \\
\cmidrule(lr){3-4}\cmidrule(lr){5-6}\cmidrule(lr){7-8}\cmidrule(lr){9-10}\cmidrule(lr){11-12}\cmidrule(lr){13-14}
 & & OSSM & SSM & OSSM & SSM & OSSM & SSM & OSSM & SSM & OSSM & SSM & OSSM & SSM \\
\midrule
\endfirsthead
\multicolumn{14}{l}{\itshape Table \ref{tab:raw-cfg7-mae}, continued.}\\
\toprule
Dataset & $H$ & \multicolumn{2}{c}{2021} & \multicolumn{2}{c}{2022} & \multicolumn{2}{c}{2023} & \multicolumn{2}{c}{2024} & \multicolumn{2}{c}{2025} & \multicolumn{2}{c}{2026} \\
\cmidrule(lr){3-4}\cmidrule(lr){5-6}\cmidrule(lr){7-8}\cmidrule(lr){9-10}\cmidrule(lr){11-12}\cmidrule(lr){13-14}
 & & OSSM & SSM & OSSM & SSM & OSSM & SSM & OSSM & SSM & OSSM & SSM & OSSM & SSM \\
\midrule
\endhead
\endfoot
\endlastfoot
\multicolumn{14}{l}{\textit{Short-term forecasting}} \\*
\addlinespace[2pt]
NN5 & 24 & \textbf{0.635} & 0.718 & \textbf{0.652} & 0.764 & \textbf{0.686} & 0.790 & \textbf{0.633} & 0.754 & \textbf{0.678} & 0.729 & \textbf{0.657} & 0.761 \\*
 & 36 & \textbf{0.627} & 0.710 & \textbf{0.659} & 0.775 & \textbf{0.698} & 0.787 & \textbf{0.631} & 0.743 & \textbf{0.674} & 0.752 & \textbf{0.673} & 0.756 \\*
 & 48 & \textbf{0.624} & 0.868 & \textbf{0.662} & 0.793 & \textbf{0.687} & 0.780 & \textbf{0.628} & 0.750 & \textbf{0.668} & 0.773 & \textbf{0.665} & 0.896 \\*
 & 60 & \textbf{0.612} & 0.705 & \textbf{0.673} & 0.772 & \textbf{0.685} & 0.806 & \textbf{0.657} & 0.803 & \textbf{0.664} & 0.768 & \textbf{0.668} & 0.899 \\
\midrule
NYSE & 24 & 0.363 & \textbf{0.344} & \textbf{0.335} & 0.337 & \textbf{0.341} & 0.382 & 0.334 & \textbf{0.328} & 0.375 & \textbf{0.354} & \textbf{0.330} & 0.698 \\*
 & 36 & 0.504 & \textbf{0.461} & 0.537 & \textbf{0.513} & \textbf{0.449} & 0.542 & \textbf{0.500} & 0.686 & \textbf{0.540} & 0.832 & \textbf{0.651} & 0.719 \\*
 & 48 & 0.693 & \textbf{0.657} & \textbf{0.762} & 0.994 & \textbf{0.671} & 0.951 & \textbf{0.716} & 0.727 & \textbf{0.623} & 0.971 & \textbf{0.462} & 1.013 \\
\bottomrule
\end{longtable}
\endgroup


\subsection{Configuration 8}

\begin{table}[H]
\centering
\caption{Hyper-parameters of configuration 8. Every other setting follows the
TFB defaults, and the two heads share all of them, including the random seeds.}
\small
\begin{tabular}{lc}
\toprule
Hyper-parameter & Value \\
\midrule
Encoder layers $e_{\mathrm{layers}}$ & 4 \\
State dimension $d_x$ & 128 \\
Model dimension $d_{\mathrm{model}}$ & 128 \\
Learning rate & $10^{-3}$ \\
Batch size & 32 \\
\bottomrule
\end{tabular}
\end{table}

\begingroup
\setlength{\tabcolsep}{2pt}
\footnotesize
\begin{longtable}{l r cc cc cc cc cc cc }
\caption{normalized MSE for configuration 8, lower is better. Rows are (dataset, forecast horizon $H$) pairs, columns are the six random seeds, each split into the observer head (OSSM) and the standard head (SSM). The better of the two is in bold. ``div.'' marks a run that diverged, i.e.\ reached a normalized MSE above 10 or produced no metric at all; divergence counts as a loss, so the head that converged is shown in bold.}\label{tab:raw-cfg8-mse}\\
\toprule
Dataset & $H$ & \multicolumn{2}{c}{2021} & \multicolumn{2}{c}{2022} & \multicolumn{2}{c}{2023} & \multicolumn{2}{c}{2024} & \multicolumn{2}{c}{2025} & \multicolumn{2}{c}{2026} \\
\cmidrule(lr){3-4}\cmidrule(lr){5-6}\cmidrule(lr){7-8}\cmidrule(lr){9-10}\cmidrule(lr){11-12}\cmidrule(lr){13-14}
 & & OSSM & SSM & OSSM & SSM & OSSM & SSM & OSSM & SSM & OSSM & SSM & OSSM & SSM \\
\midrule
\endfirsthead
\multicolumn{14}{l}{\itshape Table \ref{tab:raw-cfg8-mse}, continued.}\\
\toprule
Dataset & $H$ & \multicolumn{2}{c}{2021} & \multicolumn{2}{c}{2022} & \multicolumn{2}{c}{2023} & \multicolumn{2}{c}{2024} & \multicolumn{2}{c}{2025} & \multicolumn{2}{c}{2026} \\
\cmidrule(lr){3-4}\cmidrule(lr){5-6}\cmidrule(lr){7-8}\cmidrule(lr){9-10}\cmidrule(lr){11-12}\cmidrule(lr){13-14}
 & & OSSM & SSM & OSSM & SSM & OSSM & SSM & OSSM & SSM & OSSM & SSM & OSSM & SSM \\
\midrule
\endhead
\endfoot
\endlastfoot
\multicolumn{14}{l}{\textit{Short-term forecasting}} \\*
\addlinespace[2pt]
NN5 & 24 & 0.882 & \textbf{0.879} & \textbf{0.907} & 1.198 & 0.876 & \textbf{0.843} & 0.893 & \textbf{0.866} & \textbf{0.834} & 0.898 & \textbf{0.830} & 0.876 \\*
 & 36 & \textbf{0.929} & 1.268 & 0.900 & \textbf{0.840} & \textbf{1.296} & 1.300 & \textbf{1.303} & 1.304 & \textbf{1.089} & 1.186 & \textbf{1.108} & 1.203 \\*
 & 48 & \textbf{0.823} & 1.310 & 1.230 & \textbf{0.925} & \textbf{1.291} & 1.293 & \textbf{0.989} & 1.325 & \textbf{0.734} & 1.287 & \textbf{0.836} & 1.271 \\*
 & 60 & \textbf{1.252} & 1.297 & \textbf{1.174} & 1.219 & \textbf{1.248} & 1.285 & \textbf{0.865} & 1.308 & \textbf{0.796} & 1.220 & \textbf{0.803} & 1.191 \\
\midrule
NYSE & 24 & 0.521 & \textbf{0.331} & \textbf{0.247} & 0.492 & \textbf{0.250} & 0.292 & \textbf{0.644} & 1.189 & \textbf{0.231} & 0.250 & \textbf{0.235} & 0.253 \\*
 & 36 & \textbf{0.454} & 1.569 & \textbf{0.364} & 1.358 & 1.217 & \textbf{0.501} & \textbf{0.510} & 1.526 & \textbf{0.459} & 0.471 & \textbf{0.439} & 1.452 \\*
 & 48 & \textbf{0.685} & 1.681 & \textbf{1.568} & 1.655 & \textbf{0.708} & 1.702 & 1.727 & \textbf{1.616} & \textbf{1.459} & 1.806 & \textbf{0.748} & 1.940 \\
\bottomrule
\end{longtable}
\endgroup


\begingroup
\setlength{\tabcolsep}{2pt}
\footnotesize
\begin{longtable}{l r cc cc cc cc cc cc }
\caption{normalized MAE for configuration 8, lower is better. Rows are (dataset, forecast horizon $H$) pairs, columns are the six random seeds, each split into the observer head (OSSM) and the standard head (SSM). The better of the two is in bold. ``div.'' marks a run that diverged, i.e.\ reached a normalized MSE above 10 or produced no metric at all; divergence counts as a loss, so the head that converged is shown in bold.}\label{tab:raw-cfg8-mae}\\
\toprule
Dataset & $H$ & \multicolumn{2}{c}{2021} & \multicolumn{2}{c}{2022} & \multicolumn{2}{c}{2023} & \multicolumn{2}{c}{2024} & \multicolumn{2}{c}{2025} & \multicolumn{2}{c}{2026} \\
\cmidrule(lr){3-4}\cmidrule(lr){5-6}\cmidrule(lr){7-8}\cmidrule(lr){9-10}\cmidrule(lr){11-12}\cmidrule(lr){13-14}
 & & OSSM & SSM & OSSM & SSM & OSSM & SSM & OSSM & SSM & OSSM & SSM & OSSM & SSM \\
\midrule
\endfirsthead
\multicolumn{14}{l}{\itshape Table \ref{tab:raw-cfg8-mae}, continued.}\\
\toprule
Dataset & $H$ & \multicolumn{2}{c}{2021} & \multicolumn{2}{c}{2022} & \multicolumn{2}{c}{2023} & \multicolumn{2}{c}{2024} & \multicolumn{2}{c}{2025} & \multicolumn{2}{c}{2026} \\
\cmidrule(lr){3-4}\cmidrule(lr){5-6}\cmidrule(lr){7-8}\cmidrule(lr){9-10}\cmidrule(lr){11-12}\cmidrule(lr){13-14}
 & & OSSM & SSM & OSSM & SSM & OSSM & SSM & OSSM & SSM & OSSM & SSM & OSSM & SSM \\
\midrule
\endhead
\endfoot
\endlastfoot
\multicolumn{14}{l}{\textit{Short-term forecasting}} \\*
\addlinespace[2pt]
NN5 & 24 & 0.683 & \textbf{0.673} & \textbf{0.696} & 0.862 & 0.680 & \textbf{0.650} & 0.666 & \textbf{0.660} & \textbf{0.643} & 0.681 & \textbf{0.636} & 0.668 \\*
 & 36 & \textbf{0.727} & 0.913 & 0.711 & \textbf{0.660} & \textbf{0.928} & 0.930 & 0.930 & \textbf{0.929} & \textbf{0.815} & 0.866 & \textbf{0.833} & 0.859 \\*
 & 48 & \textbf{0.676} & 0.932 & 0.898 & \textbf{0.713} & 0.930 & \textbf{0.928} & \textbf{0.762} & 0.936 & \textbf{0.611} & 0.923 & \textbf{0.687} & 0.914 \\*
 & 60 & \textbf{0.910} & 0.930 & \textbf{0.862} & 0.889 & \textbf{0.908} & 0.927 & \textbf{0.696} & 0.933 & \textbf{0.655} & 0.888 & \textbf{0.633} & 0.878 \\
\midrule
NYSE & 24 & 0.516 & \textbf{0.410} & \textbf{0.363} & 0.526 & \textbf{0.354} & 0.377 & \textbf{0.584} & 0.862 & \textbf{0.329} & 0.339 & \textbf{0.346} & 0.350 \\*
 & 36 & \textbf{0.504} & 1.001 & \textbf{0.429} & 0.920 & 0.850 & \textbf{0.509} & \textbf{0.530} & 0.992 & 0.504 & \textbf{0.483} & \textbf{0.482} & 0.952 \\*
 & 48 & \textbf{0.638} & 1.040 & \textbf{0.991} & 1.035 & \textbf{0.645} & 1.053 & 1.061 & \textbf{1.014} & \textbf{0.952} & 1.082 & \textbf{0.658} & 1.123 \\
\bottomrule
\end{longtable}
\endgroup


\end{document}